\documentclass{article}
\pdfoutput=1 

\PassOptionsToPackage{numbers,sort&compress}{natbib}

\usepackage[preprint]{neurips_2026}

\usepackage[utf8]{inputenc}
\usepackage[T1]{fontenc}

\usepackage{hyperref}
\usepackage{url}
\usepackage{booktabs}
\usepackage{amsfonts}
\usepackage{amsmath}
\usepackage{amssymb}
\usepackage{nicefrac}
\usepackage{microtype}
\usepackage{xcolor}
\usepackage{graphicx}
\usepackage{subcaption}
\usepackage{multirow}
\usepackage{array}
\usepackage{makecell}
\usepackage{enumitem}
\usepackage{algorithm}
\usepackage{algpseudocode}
\usepackage{placeins}

\newcommand{\method}{SheafStain}

\newcommand{\eg}{\textit{e.g.}}

\newcommand{\sheafF}{\mathcal{F}}

\newcommand{\lsheaf}{\mathcal{L}_{\text{sheaf}}}

\title{SheafStain: Sheaf-Theoretic Schr\"odinger Bridge for Spatially and Biologically Coherent Virtual Staining}

\author{%
  Hyeongyeol Lim$^{1,2}$ \quad
  Hongjun Yoon$^{2}$\thanks{Corresponding authors.} \quad
  Eunjin Jang$^{2}$ \quad \\[4pt]
  \bfseries Daeky Jeong$^{2}$ \quad
  Won June Cho$^{2}$ \quad
  Hwamin Lee$^{1}$\footnotemark[\value{footnote}] \\[6pt]
  $^{1}$Department of Biomedical Informatics, Korea University College of Medicine \\
  $^{2}$DEEPNOID Inc. \\[4pt]
  \texttt{\{doodleima, hwamin\}@korea.ac.kr} \\
  \texttt{\{hylim, hyoon, ejjang, dkjeong, wjcho\}@deepnoid.com}
}

\begin{document}
\maketitle

\begin{abstract}
Current virtual staining approaches offer the potential for time- and cost-efficient 
biomarker quantification in cancer diagnostics and prognostics. 
However, patch-wise inference for gigapixel whole-slide images fails to maintain spatial continuity, 
yielding artifacts that cause catastrophic mismatches with ground-truth images.
Although pathology Vision Foundation Models (VFMs) offer rich representations, 
their self-attention causes varying global contexts to produce inconsistent embeddings 
for the same physical region. 
We formalize and validate this ``context contamination'' as a sheaf-theoretic problem 
where these embeddings form a presheaf whose sections disagree on overlaps, so no global section restricts to them.
To address this, we propose SheafStain, a new approach that reinterprets 
VFM features as sheaf-like sections for spatially and biologically coherent virtual staining. 
Specifically, SheafStain integrates class and patch tokens into a Schr\"odinger Bridge framework 
as sheaf-like sections. 
While the class token anchors biological consistency, patch tokens form a per-position spatial map.
A feature space from a single VFM encoder pretrained on Hematoxylin \& Eosin and Immunohistochemistry 
supervises both input conditioning and output stain alignment.
Departing from prior work that evaluates on isolated $256 \times 256$ patches 
and either random-crops or resizes the $1024 \times 1024$ ground truth, 
we translate at $256 \times 256$ and evaluate on the stitched $1024 \times 1024$ outputs 
across HER2, ER, PR, and Ki-67. SheafStain demonstrates promising results against six prior methods
while mitigating patch-boundary stitching artifacts. Code is available at \url{https://github.com/deepnoid-ai/SheafStain}.
\end{abstract}

\section{Introduction}
\label{sec:intro}

\begin{figure}[t]
  \centering
  \includegraphics[width=0.95\linewidth]{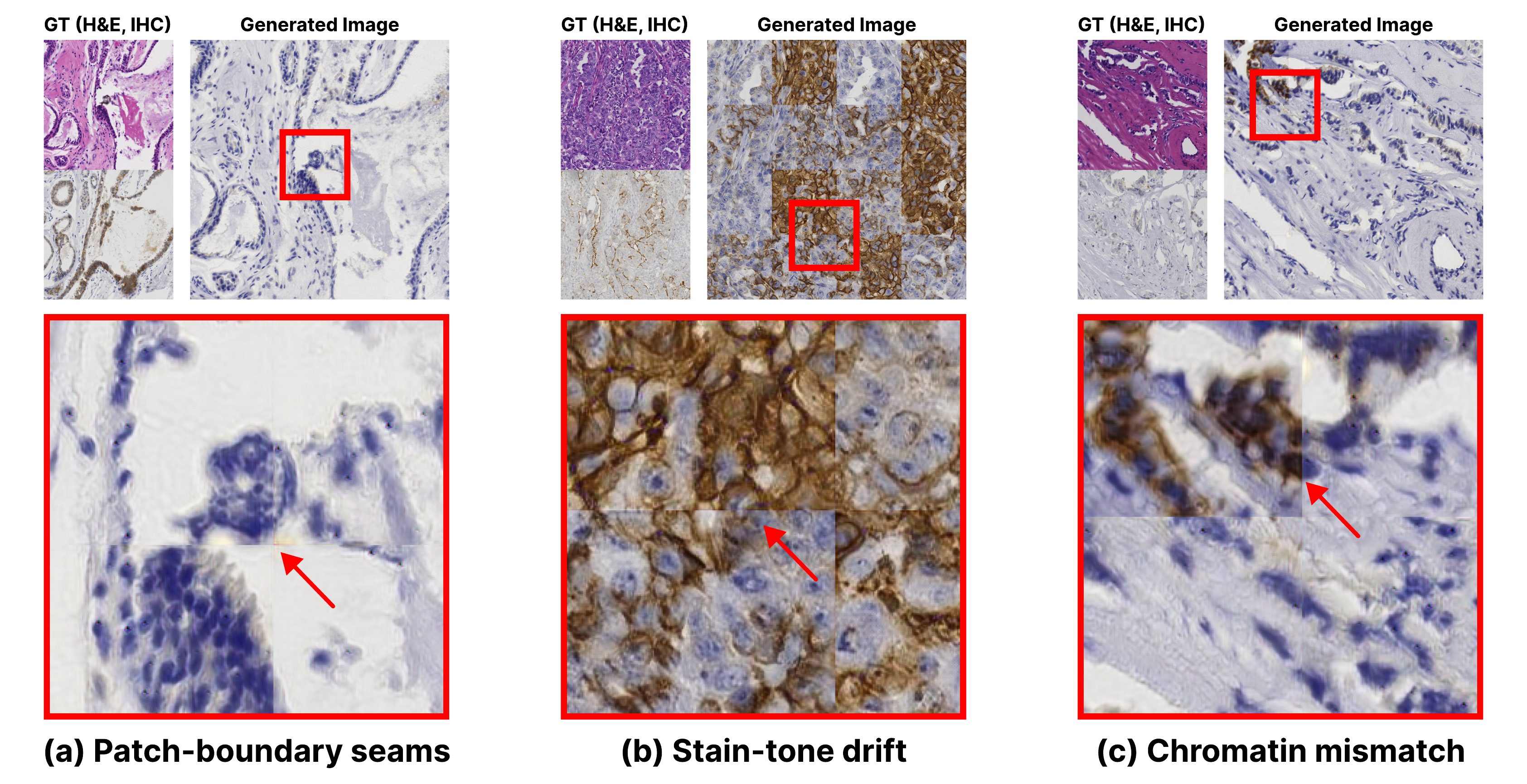}
  \caption{\textbf{Failure cases of prior virtual staining.} 
                   Independent patch translation induces
                   (a) patch-boundary discontinuities, 
                   (b) stain-tone drift, 
                   and (c) chromatin-distribution mismatch
                   across reassembled regions, disrupting morphological cues
                   used for biomarker scoring.
                   The three panels show one failure becoming visible in three ways, and they
                   can appear together in the same region. What separates them is where each acts:
                   the boundary discontinuity is the geometric signature at the seam, 
                   tone drift is a property of a whole patch relative to its neighbors, 
                   and chromatin mismatch is a texture property within the patch.}
  \label{fig:failure}
\end{figure}

Immunohistochemistry (IHC) staining is central to pathological biomarker diagnostics 
(\eg, HER2, ER/PR, Ki-67). However, only a single stain can be applied to a 
given tissue section in practice, imposing substantial resource and turnaround burdens. 
To alleviate this, a wide range of \emph{virtual staining} techniques have been actively
studied, aiming to predict the appearance of an IHC stain from a Hematoxylin \& Eosin (H\&E) image of the same
tissue so that a single section can yield several stains computationally.
Yet current approaches face three intertwined challenges.
\textbf{(i) Weakly-paired data} cannot be eliminated, because a single stain can be applied per
slice, so H\&E--IHC pairs in existing datasets come from different slices~\citep{liu2022bci,mist}.
Datasets such as HER2match~\citep{her2match} re-stain the same slice, but tissue damage and
registration drift from that process remain unresolved.
\textbf{(ii) Patch-wise isolation} compounds this, because whole-slide images (WSIs) force
independent generation on small patches (\eg, $256 \times 256$, $512 \times 512$) due to their gigapixel scale and GPU memory limits, 
so each patch is produced without access to neighboring tissue context and adjacent outputs disagree.
\textbf{(iii) Reassembled-region quality degradation} then follows, because those disagreements
translate into stitching artifacts and tone discontinuities that remain the largest barrier to
the clinical adoption of virtual staining (Figure~\ref{fig:failure}).

A natural diagnosis attributes these failures to what we term \emph{context contamination}: 
when a model generates each patch independently from a fixed-size field of view,
it lacks awareness of neighboring tissue context, so adjacent outputs disagree on stain intensity, 
nuclear texture, and structural continuity. Concretely, for a token $t$ in the overlap region $U_{ij}$ 
between two adjacent patches $U_i$ and $U_j$, a Vision Foundation Model (VFM) built on a Vision Transformer (ViT) yields the contextual representation $f_i(t)$:
\begin{equation}
  f_i(t) = \sum_{s \in U_i} \alpha^i_{ts} V_s
         = \underbrace{\sum_{s \in U_{ij}} \alpha^i_{ts} V_s}_{\text{shared content}}
         + \underbrace{\sum_{s \in U_i \setminus U_{ij}} \alpha^i_{ts} V_s}_{\text{contamination}},
  \label{eq:context_full}
\end{equation}
where $\alpha^i_{ts}$ are the attention weights and $V_s$ are the value vectors. 
Because the same token in $U_j$ yields a different $f_j(t)$ via different non-overlapping 
context, the shared physical tissue receives inconsistent representations across patches. 

We argue that a more fundamental remedy must intervene earlier, 
by injecting \emph{neighborhood-aware context} directly into the generative process 
so that each patch is synthesized in agreement with its surrounding tissue.
From a sheaf-theoretic perspective, an aggregated histopathology image is 
naturally modeled as a presheaf of local sections over an open cover of $X$ by overlapping patches, 
and stitching consistency reduces to enforcing the compatibility that gluing requires on overlaps.
For virtual staining, the output is consumed as an assembled image. A disagreement between
two patches on their overlap appears as the failures in Figure~\ref{fig:failure}.
Furthermore, this representational mismatch is most pronounced at the \emph{two extremes}
of an image's frequency-content distribution: homogeneous background and densely structured tissue regions.

In this work, we present SheafStain as a framework that operationalizes this perspective. 
SheafStain exploits the practically unpaired nature of weakly-paired H\&E--IHC datasets, 
and enforces sheaf-theoretic consistency across both training and inference so that structural coherence 
is maintained from patch boundaries up to larger reassembled regions. Our contributions are summarized as follows:

\begin{enumerate}[leftmargin=*,itemsep=2pt]
\item \textbf{VFM features as sheaf-structured spatial conditioning within a Schr\"odinger Bridge (SB) framework:}
              To our knowledge, SheafStain is the first virtual staining framework that interprets pathology
              VFM tokens as sheaf sections over an open cover and feeds them as a per-position spatial map
              (rather than a single global vector) within an SB framework, enabling principled
              local-to-global conditioning.

  \item \textbf{Sheaf-consistent training and inference:} 
                At training time, we introduce a \emph{pixel sheaf loss} and 
                a \emph{cocycle gluing loss} that explicitly enforce structural consistency 
                between adjacent patches. At inference time, we preserve the same sheaf-consistent 
                conditioning procedure used during training, and additionally adopt an inference strategy 
                that adaptively places extra patches at the two extremes of the frequency-content 
                distribution to refine the open cover, thereby mitigating stitching artifacts.

\item \textbf{Empirical validation at the resolution that matters:}
              Prior virtual staining work evaluates almost exclusively on isolated $256 \times 256$ patches.
              Standard random-cropping or resizing protocols mask the very patch-boundary artifacts that
              determine clinical usability. We translate at $256 \times 256$ and stitch back to the native
              $1024 \times 1024$ of BCI and MIST datasets, evaluating \emph{on the stitched output} with a
              comprehensive metric suite (Section~\ref{sec:setup}).
              We compare SheafStain against six prior methods~\citep{pix2pix,cyclegan,pspstain,dvst,unistainnet,unsb}
              across HER2, ER, PR, and Ki-67.
\end{enumerate}

\section{Related Work}
\label{sec:related}

\paragraph{Generative approaches for virtual staining.}
Translation between H\&E and IHC has been pursued across architectural families. 
Paired and unpaired GAN approaches~\citep{pix2pix,cyclegan,pspstain,liu2022bci,mist,structural_cyclegan}
established the standard pipelines for cross-stain image translation.
In particular, CC-WSI-Net~\citep{liu2024generating} addresses patch-stitching artifacts in WSI synthesis via dedicated content and color consistency modules,
supervising tile borders during training and assembling the slide from center crops at inference,
and ODA-GAN~\citep{wang2025odagan} combines weakly-supervised segmentation with orthogonal feature decoupling to separate stain-related from stain-unrelated factors during H\&E$\to$IHC translation.
Diffusion-based methods~\citep{histdit,staindiffuser} have since replaced
adversarial training with iterative denoising for improved fidelity. 
D-VST~\citep{dvst} employs a diffusion transformer to decouple tone control 
from pathology via curriculum learning, while introducing a frequency-aware 
sampling strategy to mitigate mosaic artifacts during inference.
SB-based approaches reframe translation as an entropy-regularized
stochastic interpolation between two distributions. 
Unpaired Neural Schr\"odinger Bridge (UNSB)~\citep{unsb} is a representative neural realization that achieves competitive unpaired translation
by iterative adversarial decomposition, and PASB~\citep{pasb} extends this line with pathology-aware 
supervision. 
Building on UNSB as our starting point, SheafStain treats the inconsistency in the VFM
presheaf as a failure of the compatibility that gluing requires on overlaps, and introduces
sheaf-theoretic conditioning and consistency losses that constrain those overlaps directly.

\paragraph{Sheaf theory in deep learning.}
Sheaf theory has recently been introduced into deep learning, primarily for graph and hypergraph neural networks. 
Neural Sheaf Diffusion (NSD)~\citep{nsd} equips graphs with non-trivial cellular sheaves 
and shows that learnable restriction maps mitigate oversmoothing and heterophily. 
Sheaf Hypergraph Networks (SHN)~\citep{shn} extend this to hypergraphs with diagonal
and low-rank parameterizations, while Learning Sheaf Laplacian (LSL)~\citep{lsl} infers
graph topology and restriction maps from node-observed data via closed-form
total-variation minimization.
Cooperative Sheaf Neural Networks (CSNN)~\citep{csnn} further extend the family of
sheaf-based GNNs. All these works operate on discrete graph structures. 
To the best of our knowledge, SheafStain is the first to apply \textbf{sheaf theory} to image generation, 
specifically for enforcing local-to-global spatial consistency within a Schr\"odinger Bridge framework.

\section{Method}
\label{sec:method}

\subsection{Preliminaries}
\label{sec:prelim}

SheafStain combines three components. 
\textbf{Sheaf theory}~\citep{bredon1997sheaf,sheaf_survey} formalizes spatial coherence. 
A \emph{presheaf of vector spaces} $\sheafF$ on $X$ assigns to each open $U \subseteq X$ 
a vector space $\sheafF(U)$ together with restriction maps $\rho_{U \to V}: \sheafF(U) \to \sheafF(V)$ 
for $V \subseteq U$. 
It becomes a \emph{sheaf} when locally compatible sections on an open cover glue 
uniquely to a global section. Deviation from the compatibility that gluing requires is
measured by the sheaf Laplacian $\Delta_0 = \delta^*\delta$ 
with Dirichlet energy $E(x) = \langle \Delta_0 x, x \rangle \ge 0$~\citep{nsd}. 
\textbf{Schr\"odinger Bridge (SB)}~\citep{schrodinger1932, leonard2013sb} 
solves $Q_{\mathrm{SB}} = \arg\min_{Q \in \mathcal{D}(\pi_0,\pi_1)} \mathrm{KL}(Q \,\|\, W^\tau)$, 
where KL is the Kullback--Leibler divergence, recovering Optimal Transport (OT)~\citep{villani2009ot} as $\tau \to 0$~\citep{chen2021sbotconnection}.
We adopt this formulation for two reasons. First, virtual staining takes H\&E as the source
and IHC as the target, which places it close to image-to-image translation. Both endpoint
distributions are therefore observed, and the bridge connects them directly, whereas a diffusion
path takes a Gaussian prior as its starting point instead of the source distribution.
Second, most virtual staining datasets are curated from consecutive sections of the same tissue,
so the two stains are only weakly paired. UNSB~\citep{unsb} learns this source-to-target bridge
from unpaired data instead of pixel-level pairs, and we therefore build on it.
\textbf{Pathology vision foundation model}~\citep{gigapath} (Prov-GigaPath, co-pretrained on 1.3B H\&E and IHC tiles from 171K slides)
yields sections for both stains in one feature space (Appendix~\ref{app:gluing}),
on which the restriction maps used in Section~\ref{sec:sheaf} operate.
Formal definitions (presheaf, gluing axiom, cocycle condition) are presented in Appendix~\ref{app:sheaf}. 
The SB$\leftrightarrow$OT correspondence, UNSB's adversarial decomposition, and SheafStain's regularized-SB positioning 
are detailed in Appendix~\ref{app:theory}.

\paragraph{Notation.}
$X$: image domain; 
$\mathcal{U}=\{U_i\}$: open cover of $X$ by overlapping patches; 
$U_{ij}=U_i\cap U_j$. $\Phi$: frozen VFM; 
$s_i = \Phi(U_i) \in \sheafF(U_i)$: VFM section over $U_i$ ($T{=}196$ tokens, each in
$\mathbb{R}^{d}$, where $d$ is the embedding dimension of the VFM encoder, \eg, $1536$ for
Prov-GigaPath).
$\rho_{U_i \to U_{ij}}$: restriction by spatial token indexing. 
$\delta, \Delta_0, E(\cdot)$: coboundary, sheaf Laplacian, Dirichlet energy. 
$\pi_0,\pi_1$: H\&E and IHC marginals; $\tau$: SB diffusion coefficient. 
$G$: SB generator; $\lambda_\bullet$: loss weights (introduced with each loss term).

\subsection{Sheaf-Theoretic Problem Formulation}
\label{sec:sheaf}

Let $X$ denote a histopathology image 
and $\mathcal{U}=\{U_i\}$ an open cover of $X$ by overlapping patches. 
We introduce the vocabulary on one pair of them, laid out in Table~\ref{tab:worked_pair}: two
adjacent patches at the stride-$192$ setting of Appendix~\ref{app:gluing}, meeting in a strip
$U_{12} = U_1 \cap U_2$. The frozen VFM tokenizes each patch into a $14 \times 14$ grid of
$16 \times 16$ px cells that cover it, one token per cell.

To a patch $U$ we attach $\sheafF(U)$, the vector space of all assignments of $d$-dimensional
vectors to the token positions of $U$. The assignment the VFM actually produces,
$s_i := \Phi(U_i) \in \sheafF(U_i)$, is a \emph{section over} $U_i$. The restriction
$\rho_{U_i \to U_{12}}$ keeps the $2 \times 14$ tokens whose cells lie in the strip, so both readings land in
the same $28$-token space. Restriction is positional indexing, so restricting in one step and
restricting through an intermediate region return the same tokens (max difference $= 0$), and
the assignment is a presheaf on $X$. Of the two sheaf axioms, \emph{locality} is immediate here,
since a section is determined by its token positions.

\begin{table}[h]
  \centering
  \caption{\textbf{Example pair of overlapping patches.} The sheaf vocabulary instantiated on one
  pair at stride $192$. $\rho$ denotes positional token selection.}
  \label{tab:worked_pair}
  \footnotesize
  \begin{tabular}{lccc}
    \toprule
     & $U_1$ & $U_2$ & $U_{12}$ \\
    \midrule
    anchor (px)        & $(0,0)$              & $(192,0)$            & $(192,0)$ \\
    size (px)          & $224 \times 224$     & $224 \times 224$     & $32 \times 224$ \\
    tokens             & $14 \times 14 = 196$ & $14 \times 14 = 196$ & $2 \times 14 = 28$ \\
    selected by $\rho$ & rightmost 2 columns  & leftmost 2 columns   & all columns \\
    \bottomrule
  \end{tabular}
\end{table}

\emph{Compatibility} is what fails. The strip is one piece of tissue read twice, once as the
right edge of $U_1$ and once as the left edge of $U_2$, so gluing requires
$\rho_{U_1 \to U_{12}}(s_1) = \rho_{U_2 \to U_{12}}(s_2)$. Over $4{,}873$ BCI
images~\citep{liu2022bci} the mean cosine between the two readings is $0.63$ at this stride and
reaches only $0.92$ at $86\%$ overlap, never the $1.0$ the condition demands. No single section
over $U_1 \cup U_2$ can restrict to both, since its restrictions would agree on the strip by the
definition of $\rho$. Summed over every overlap in the cover, these squared disagreements are
the Dirichlet energy $E(x)$ of Section~\ref{sec:prelim}, which is therefore strictly positive.
This is the inconsistency that manifests as stitching artifacts under patch-wise translation.
The mechanism is \emph{context contamination} in global self-attention,
where the same token at position $t$ aggregates different non-overlapping context
in $U_i$ vs.\ $U_j$. The near-identical gap magnitude across H\&E and IHC
($\Delta < 0.02$) confirms stain-independence.
The implication for patch-wise SB is that translated patches inherit the presheaf structure 
and remain locally valid but globally incoherent. 
Rather than post-hoc VFM correction, SheafStain injects spatial context
as conditioning and enforces the compatibility that gluing requires via explicit losses, 
letting the generator learn \emph{inherent restriction maps}
for global consistency (Appendix~\ref{app:training}).
Sheaf constructions in deep learning are reviewed in Section~\ref{sec:related}, the formal
definitions are collected in Appendix~\ref{app:sheaf}, and full empirical verification, together
with a discussion of which encoders context contamination applies to, is in
Appendix~\ref{app:gluing}.

\subsection{Pipeline Overview}
\label{sec:pipeline}

SheafStain instantiates the sheaf framework as a pipeline symmetric 
across training and inference: in both phases the spatial map 
and the neighborhood class token (CLS token) are extracted from a 
single VFM forward pass and serve as the generator's conditioning.

\begin{figure}[t]
  \centering
  \includegraphics[width=\linewidth]{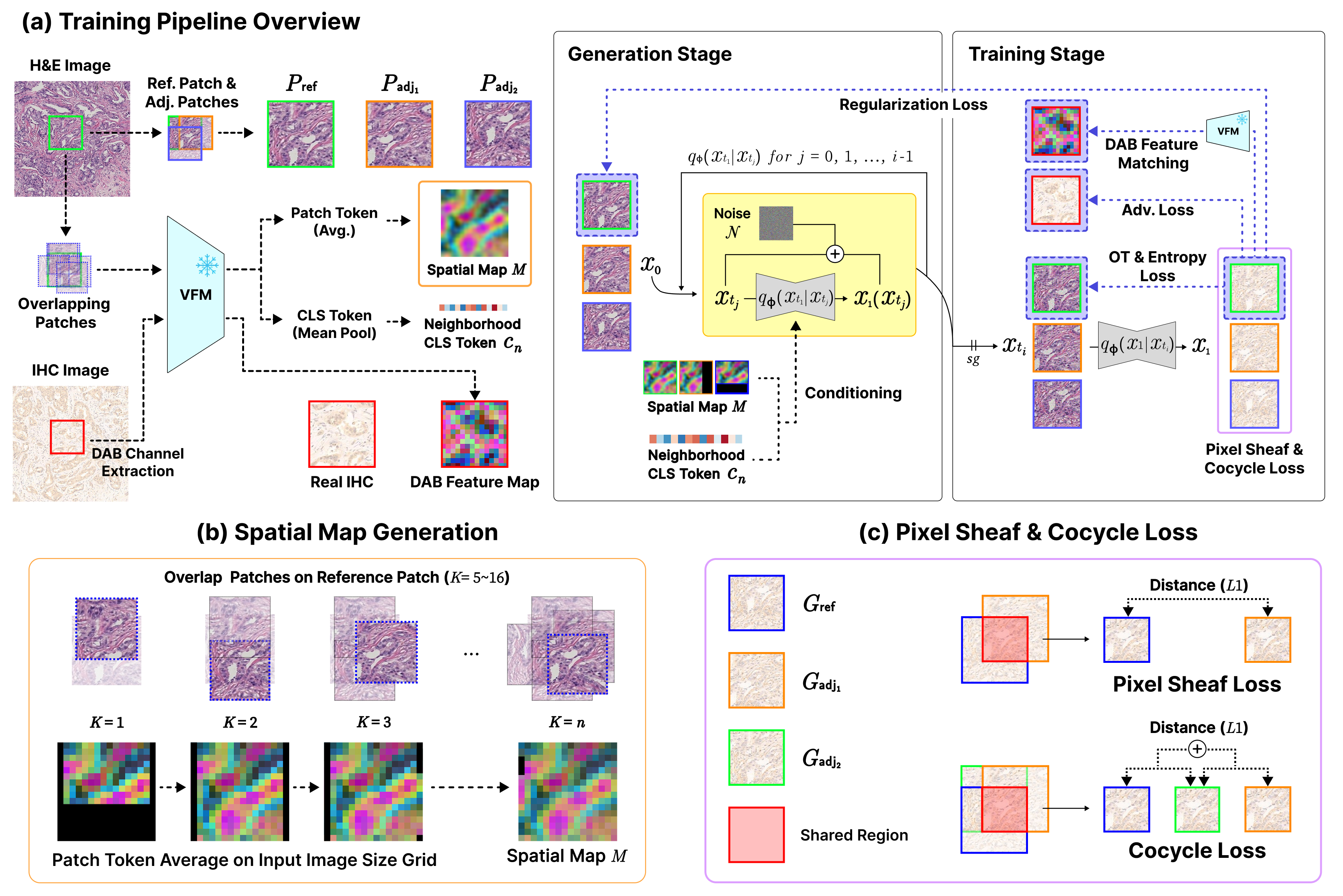}
  \caption{\textbf{SheafStain training pipeline.}
  \textbf{(a) Overview.} Reference patch $P_{\text{ref}}$ and two adjacent
  targets $P_{\text{adj}_1}, P_{\text{adj}_2}$ sharing a triple-overlap are
  sampled. VFM-derived $M$ and $c_n$ condition three weight-shared generator
  passes that yield $G_{\text{ref}}, G_{\text{adj}_1}, G_{\text{adj}_2}$.
  \textbf{(b) Spatial map.} VFM tokens of overlapping patches are aggregated
  into $M$, a sheaf-consistent section over $P_{\text{ref}}$.
  \textbf{(c) Pixel sheaf and cocycle losses.} Penalize disagreement between
  adjacent outputs on shared regions, so the generator learns sheaf-consistent
  restriction maps.}
  \label{fig:flow}
\end{figure}

\paragraph{Training.}
Figure~\ref{fig:flow} illustrates the construction.
Each step samples a reference patch $P_{\text{ref}}$ and two adjacent targets $P_{\text{adj}_1}, P_{\text{adj}_2}$ 
with non-empty triple-overlap $\mathcal{O}_3$. The frozen VFM produces the spatial conditioning map $M$ 
and neighborhood CLS token $c_{n}$ once over the overlapping patches $\mathcal{P}_{\text{ref}}$. 
The spatial conditioning map $M$ is shifted to align with each adjacent frame so that conditioning agrees on $\mathcal{O}_3$ 
by construction, and $c_{n}$ is shared. Three weight-shared generator passes 
yield $G_{\text{ref}}, G_{\text{adj}_1}, G_{\text{adj}_2}$, scored by the sheaf and cocycle losses 
(Section~\ref{sec:sheaf_loss}) together with the standard baseline losses. 
The full objectives, including pathology-specific regularizers, are presented in 
Appendices~\ref{app:loss},~\ref{app:training}.

\begin{figure}[t]
  \centering
  \includegraphics[width=\linewidth]{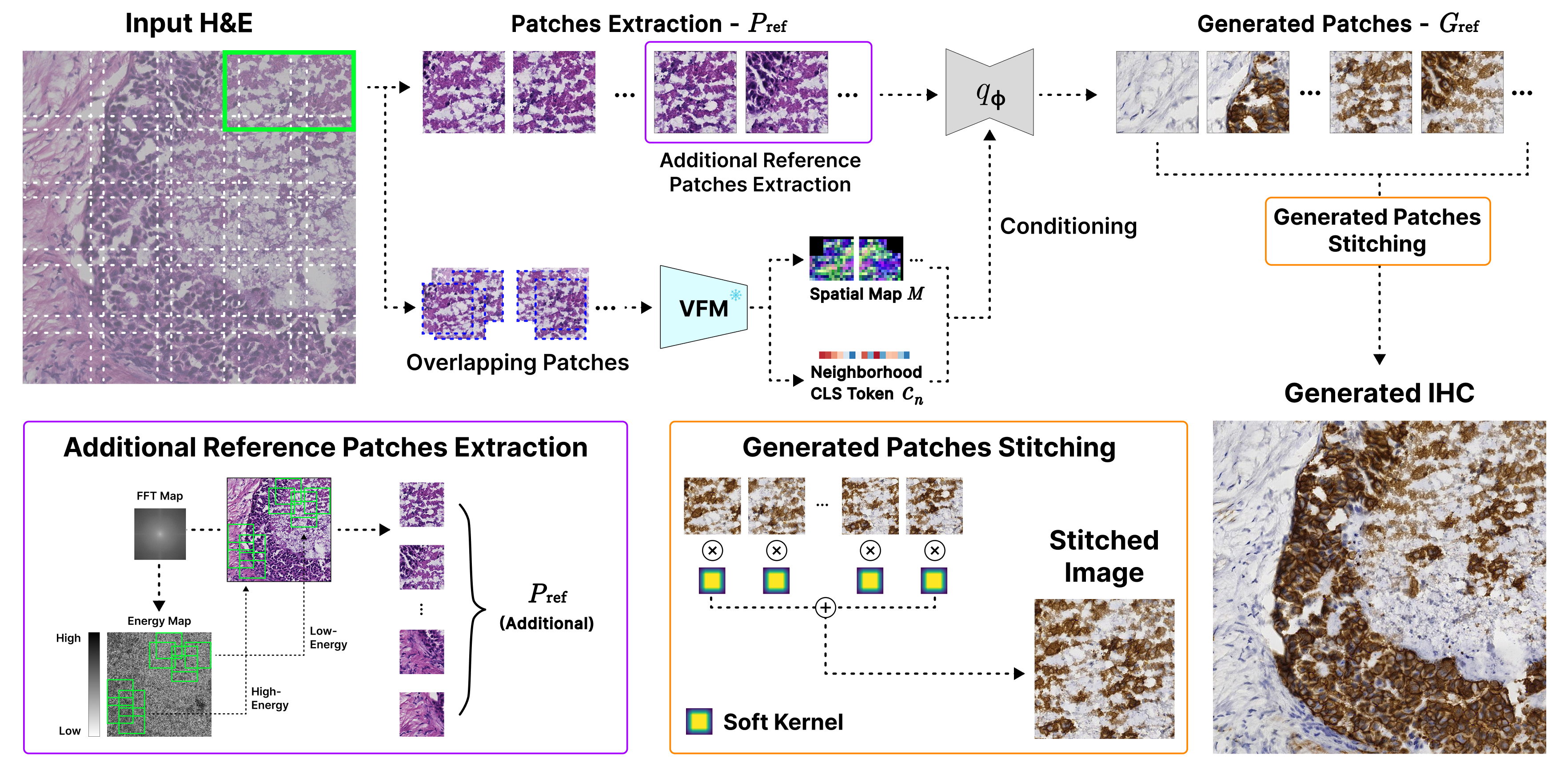}
  \caption{\textbf{SheafStain inference pipeline.} 
  The input H\&E is covered by a stride-driven reference grid with overlapping adjacent patches 
  (white guides), supplemented with additional reference patches drawn from the high- and low-energy extremes 
  of the mid-band Fast Fourier Transform (FFT) energy map (lower-left). Each reference patch is independently translated
  by the generator $q_{\phi}$ under VFM-derived conditioning ($M$, $c_{n}$), 
  and the resulting patches are reassembled by soft-kernel weighted blending (lower-right).}
  \label{fig:inference_pipeline}
\end{figure}

\paragraph{Inference.}
Figure~\ref{fig:inference_pipeline} illustrates the procedure.
At inference time, SheafStain covers the input H\&E with a $256 \times 256$ reference window
at stride $< 256$ ($192$ in this work), so that adjacent reference patches share an overlap region. 
This adjacency-by-overlap structure removes the need for a separate adjacency-sampling pass. 
To extend the regular-grid cover, we sample additional reference patches from the high- and 
low-energy extremes of the mid-band FFT energy map (preserving the open-cover randomness of the sheaf construction).
The VFM is applied once over the image. Each reference patch $P_{\text{ref}}$ obtains its
conditioning $(M, c_n)$ by spatial slicing of the global token map (realizing
$\rho_{X, P_{\text{ref}}}: \mathcal{F}(X) \to \mathcal{F}(P_{\text{ref}})$).
Each reference patch is then independently translated and patches are reassembled into the
$1024 \times 1024$ output by soft-kernel weighted blending. 
The full procedure is presented in Appendix~\ref{app:inference}.

\subsection{Spatial Map and Neighborhood Class Token as Conditioning}
\label{sec:conditioning}

\paragraph{VFM for morphological and biological context.}
SheafStain decouples the VFM's contribution into morphological context,
encoded from raw H\&E patches via the spatial conditioning ($M$, $c_{n}$),
and biological context, encoded via \emph{VFM Alignment}
(Appendix~\ref{app:cs-vfm-align}) on renderings that isolate diaminobenzidine (DAB),
the IHC chromogen channel obtained via color deconvolution~\citep{ruifrok2001}.
Each VFM embedding of an H\&E patch is a \emph{section over} that patch,
a local information container attached to the region it covers.
Adjacent sections disagree on overlaps due to context contamination,
but they still encode rich tissue-structure information that disambiguates
the H\&E$\to$IHC mapping under the weak pairing of consecutive sections.
This allocation distinguishes SheafStain from approaches~\citep{histdit}
that treat VFM embeddings as a single global biological-phenotype signal 
via conditional normalization.

\paragraph{Spatial map and neighborhood class token.}
For a reference patch $P_{\text{ref}}$ we define $\mathcal{P}_{\text{ref}}$ 
as the set of $K := |\mathcal{P}_{\text{ref}}|$ overlapping $224 \times 224$ patches 
around $P_{\text{ref}}$ ($K \in [5, 16]$, varying with tissue coverage) 
and process all of them in a single VFM forward pass, 
producing for each a $14 \times 14$ grid of patch tokens together with a CLS token.
Position-mapping and averaging of patch tokens yields the spatial conditioning map 
\begin{equation}
  M \in \mathbb{R}^{{d} \times 16 \times 16},
  \label{eq:spatial_map}
\end{equation}
where each of the $16 \times 16$ positions carries its own local embedding over $P_{\text{ref}}$. 
The CLS tokens are mean-pooled into a single neighborhood vector $c_{n} \in \mathbb{R}^{{d}}$
summarizing the surrounding context. 
Both streams are coherent, derived from the same pool of overlapping patches at the same spatial scale,
in a single VFM pass. 
During training, $(M, c_{n})$ is reused for adjacent patches by token-grid shifting $M$ 
to each adjacent frame so overlapping positions receive identical conditioning, 
while $c_{n}$ is shared. 
$M$ and $c_{n}$ are injected into each residual block of the generator 
alongside the time embedding $e_t$:
\begin{equation}
  h' = h + W_{\text{time}}(e_t) + W_{\text{spatial}}(M^{\uparrow}) + W_{\text{cls}}(c_{n}),
  \label{eq:injection}
\end{equation}
where $h$ is the intermediate feature map, $M^{\uparrow}$ is the bilinearly upsampled spatial map, 
and $W_{\text{spatial}}, W_{\text{cls}}$ are learned projections. 
Both $W_{\text{spatial}}$ and $W_{\text{cls}}$ are zero-initialized, 
ensuring conditioning starts as a no-op and grows in influence during training.
Figure~\ref{fig:pipeline} (Appendix~\ref{app:implementation}) illustrates the construction and the exact extraction geometry.

\subsection{Sheaf-Theoretic Regularizers}
\label{sec:sheaf_loss}

Each step samples $P_{\text{ref}}$ and two adjacent targets 
$P_{\text{adj}_1}, P_{\text{adj}_2}$ 
with a non-empty triple-overlap 
$\mathcal{O}_3 := U_{\text{ref}} \cap U_{\text{adj}_1} \cap U_{\text{adj}_2}$, 
distinct from the conditioning patches $\mathcal{P}_{\text{ref}}$. 
The three generator outputs $G_{\text{ref}}, G_{\text{adj}_1}, G_{\text{adj}_2}$ 
are shared across the regularizers below. 
Sampling geometry is detailed in Appendix~\ref{app:implementation}.

\paragraph{Pixel sheaf and cocycle loss.}
Spatial conditioning shares context but does not enforce output consistency.
We enforce the compatibility that gluing requires (Eq.~\ref{eq:compatibility}) on the overlap $\mathcal{O} := U_i \cap U_j$:
\begin{equation}
  \lsheaf(G_i, G_j) = \big\| \mu(G_i|_{\mathcal{O}}) - \mu(G_j|_{\mathcal{O}}) \big\|_1
                    + \alpha \big\| G_i|_{\mathcal{O}} - G_j|_{\mathcal{O}} \big\|_1,
  \label{eq:sheaf_loss}
\end{equation}
where $\mu$ is the per-channel spatial mean and $\alpha$ balances tone vs.\ structural agreement.
The mean term targets boundary tone (a common stitching-artifact source) and provides
gradient when pixel-wise disagreement is trivially small.
For global consistency, a \v{C}ech 0-cocycle requires pairwise-agreeing restrictions on
overlaps~\citep{bredon1997sheaf} (Appendix~\ref{app:sheaf}). On $\mathcal{O}_3$, we enforce this
condition across the three generated sections by applying $\lsheaf$ to all three pairs,
collecting the two not in Eq.~\ref{eq:sheaf_loss}:
\begin{equation}
  \mathcal{L}_{\text{cocycle}} = \lsheaf(G_{\text{ref}},\, G_{\text{adj}_2})
                                + \lsheaf(G_{\text{adj}_1},\, G_{\text{adj}_2}),
  \label{eq:cocycle_loss}
\end{equation}
together driving $g_{ij} \to 0$. Since the three generator passes share $c_{n}$ and
conditioning over $\mathcal{O}_3$, penalized disagreement reflects generator inconsistency,
not conditioning variation. 
Both losses compare generated patches pairwise rather than against ground truth (GT), 
making them GT-free and robust under weak GT pairing.

\section{Experiments}
\label{sec:experiments}

\subsection{Experimental Setup}
\label{sec:setup}

\paragraph{Datasets.}
We use the two standard public virtual staining datasets,
BCI~\citep{liu2022bci} and MIST~\citep{mist}, at $1024 \times 1024$ native resolution.
The choice follows from what our subject requires. Patch assembly under weak pairing needs an
IHC reference that corresponds spatially to a region wide enough to contain interior patch
boundaries, and it needs the consecutive-section regime that produces the weak pairing in the first
place. BCI and MIST both provide registered $1024 \times 1024$ pairs cut from consecutive sections.
They are also the evaluation sets this literature already reports on, so our retrained numbers can
be checked against published ones.
HER2 staining is available in both datasets.
ER, PR, and Ki-67 are available only in MIST. BCI provides an official train/test split 
accompanied by HER2 grading labels (0, 1+, 2+, 3+), whereas MIST provides an official 
train/validation split. We follow the split configuration used in prior work~\citep{unistainnet}. 
For BCI, we train on the train split and evaluate on test. For MIST, we train on the train split 
and evaluate on validation (since MIST does not provide a separate test split). 
No validation-based model selection is performed. Final-epoch weights are used for all methods. 
Per-marker pair counts are reported in Appendix~\ref{app:datasets}.

\paragraph{Training and evaluation.}
SheafStain uses a ResNet-based generator with additive sheaf conditioning. 
We train with Adam~\citep{kingma2015adam} (lr $2{\times}10^{-4}$, $\beta{=}(0.5, 0.999)$), 
an effective batch size of 192 (24 per GPU $\times$ 8 NVIDIA H200 GPUs), for 400 epochs 
(200 fixed + 200 linear decay). Prov-GigaPath is kept frozen.
Loss weights and curriculum schedule are detailed in Appendix~\ref{app:training}.
Full implementation details are in Appendix~\ref{app:implementation}.

We compare SheafStain against six prior methods: pix2pix, CycleGAN, PSPStain, D-VST, UNIStainNet, and UNSB.
We select these baselines to meet two conditions: (1) one representative from each
methodology used for virtual staining, and (2) public source code, since comparing the results
requires retraining every method on the same data. UNSB is both a baseline and our starting point,
so the comparison isolates what our components add.
We retrain each method from publicly available source code under its paper-prescribed protocol.
All methods use $256 \times 256$ patches extracted from the BCI and MIST datasets
with the official splits (train/test for BCI, train/validation for MIST).
We evaluate at the $1024 \times 1024$ aggregated image level for large-scale spatial consistency.
SheafStain and D-VST each follow their paper-prescribed inference
(SheafStain: Section~\ref{sec:pipeline}, Appendix~\ref{app:inference}).
For the other five prior methods (pix2pix, CycleGAN, PSPStain, UNSB, UNIStainNet),
we arrange $256 \times 256$ outputs $4 \times 4$ row-major into $1024 \times 1024$,
the minimal protocol consistent with their patch-level training.

\paragraph{Evaluation metrics.}
We report a comprehensive metric suite across four categories.
\emph{Distribution-level}: Fr\'echet Inception Distance (FID)~\citep{heusel2017fid} and Kernel Inception Distance (KID$\times 10^{3}$)~\citep{kid} quantify the gap 
between the generated and target IHC distributions.
\emph{Full-reference per-image}: Learned Perceptual Image Patch Similarity (LPIPS)~\citep{lpips}, Deep Image Structure and Texture Similarity (DISTS)~\citep{dists}, Structural Similarity Index Measure (SSIM)~\citep{ssim},
and Peak Signal-to-Noise Ratio (PSNR) measure perceptual and structural agreement with paired GT.
\emph{Stitching-specific}: the Tiling Score (TS)~\citep{tiled_diffusion} measures
boundary discrepancy at patch seams.
\emph{Pathology-specific}: DAB Pearson correlation (DAB-$r$), DAB KL divergence (DAB-KL), DAB Jensen--Shannon divergence (DAB-JSD),
absolute errors in mean integrated optical density (mIOD), the mean DAB optical density over pixels above $\tau = 0.1$, and fraction of optical-density positive area (FOD), the fraction of pixels above the same threshold.
Together, the stitching-specific, pathology-specific, and full-reference categories cover the
failure modes of Figure~\ref{fig:failure}.

\paragraph{HER2 binary classification.}
We further test downstream clinical utility through a HER2 grade classification task on BCI.
The dataset provides only the four IHC grades, without in situ hybridization (ISH) results or sub-categories
of IHC $0$, so HER2-ultralow within IHC $0$ and HER2-low status among $2+$ cases cannot be assigned.
We therefore collapse the grades into Absent-to-Low (HER2-AL; $0, 1+$) and Equivocal-to-Positive
(HER2-EP; $2+, 3+$). For these two groups, ASCO/CAP reporting~\citep{wolff2018asco} classifies every grade in HER2-AL as negative
and every grade in HER2-EP as equivocal ($2+$) or positive ($3+$). Negative results need no further HER2
testing when no histopathologic discordance is apparent, whereas equivocal results require reflex ISH.
The task therefore evaluates whether the outputs can serve as a screening tool that separates the two groups.
For each model, we train a ResNet-50 classifier on the method's translated train split
and evaluate on the real BCI test set.
The result accuracy on real test data reflects whether translated outputs 
preserve discriminative features that transfer to real diagnostic data.
Full protocol details are shown in Appendix~\ref{app:downstream}.

\subsection{Ablation Study and Comparison with Prior Methods}
\label{sec:ablation}
\label{sec:baselines}

We perform a progressive ablation to quantify the contribution of each SheafStain component.
Starting from the baseline, we sequentially add SheafStain's components:
(i) \emph{spatial conditioning} ($M$ and $c_n$),
(ii) \emph{pixel sheaf loss} on pairwise overlaps,
(iii) \emph{cocycle loss} extending the sheaf penalty to triple overlaps,
(iv) \emph{Fourier edge loss} for high-frequency structure preservation,
(v) \emph{DAB intensity loss} matching top-10\% chromogen intensity, and
(vi) \emph{VFM alignment} on DAB-isolated renderings, completing SheafStain.
Splitting (ii) and (iii) isolates each term's marginal contribution,
showing whether cocycle adds gains beyond pairwise compatibility.
We evaluate at $1024 \times 1024$ on the BCI test set.

Table~\ref{tab:ablation} reports the sheaf-component ablation (left) and the HER2
binary classification (right).
On the ablation, spatial conditioning produces the dominant single-step gain across all
metrics, establishing that VFM-derived tissue context resolves the bulk of patch-wise inconsistency
before any explicit constraint. Pixel sheaf and cocycle losses further tighten distribution. TS
plateaus after the pixel sheaf loss, indicating pairwise supervision saturates boundary
consistency at $1024 \times 1024$. Fourier edge loss yields the largest perceptual-quality jump
and adds a further gain in DAB-$r$. DAB-$r$ rises across the sheaf-related steps, so the
sheaf objectives improve biomarker localization on their own. DAB intensity loss and VFM
Alignment add small but consistent
gains. The full SheafStain attains the best FID, LPIPS, DISTS, and DAB-$r$.
On the classification, SheafStain leads accuracy, F1, and area under the receiver operating characteristic curve (AUROC) on real BCI test data, exceeding
the strongest prior method UNIStainNet by 0.042 in accuracy and 0.057 in AUROC.

\begin{table}[t]
  \caption{\textbf{Ablation study (left) and HER2 binary classification results (right) on BCI.} Best metrics in bold, second-best metrics underlined.}
  \label{tab:ablation}
  \label{tab:downstream_cls}
  \centering
  \footnotesize
  \setlength{\tabcolsep}{2pt}
  \resizebox{\textwidth}{!}{%
  \begin{tabular}{lcccccc@{\hspace{10pt}}|@{\hspace{10pt}}lccc}
    \toprule
    Method & FID$\downarrow$ & KID$\times10^{3}\downarrow$ & LPIPS$\downarrow$ & DISTS$\downarrow$ & TS$\downarrow$ & DAB-$r$$\uparrow$ & Method & Acc.$\uparrow$ & F1$\uparrow$ & AUROC$\uparrow$ \\
    \midrule
    Baseline (UNSB)                  & 227.7889 & 236.2310 & 0.6635 & 0.3347 & 0.1464 & 0.0589 & pix2pix & 0.660 & 0.774 & 0.625 \\
    \,\,+ Spatial Cond.                    & 55.2489 & 16.5750 & 0.5538 & 0.2960 & 0.0143 & 0.0823 & CycleGAN & 0.708 & 0.826 & 0.577 \\
    \,\,+ Pixel Sheaf $\mathcal{L}$        & 41.5356 & 6.8270 & 0.5225 & 0.2721 & \textbf{0.0131} & 0.1059 & PSPStain & 0.652 & 0.776 & 0.527 \\
    \,\,+ Cocycle $\mathcal{L}$            & 38.0140 & 4.6070 & 0.5176 & 0.2679 & \underline{0.0132} & 0.1091 & D-VST & 0.658 & 0.761 & 0.650 \\
    \,\,+ Fourier Edge $\mathcal{L}$       & \underline{36.5335} & 5.1630 & 0.4774 & 0.2285 & 0.0145 & \underline{0.1190} & UNIStainNet & \underline{0.724} & \underline{0.835} & \underline{0.737} \\
    \,\,+ DAB Intensity $\mathcal{L}$      & 37.0935 & \textbf{4.1570} & \underline{0.4718} & \underline{0.2187} & 0.0144 & 0.1183 & UNSB & 0.524 & 0.640 & 0.460 \\
    \,\,\textbf{+ VFM Align. $\mathcal{L}$\,(ours)}        & \textbf{36.3626} & \underline{4.2070} & \textbf{0.4689} & \textbf{0.2132} & 0.0146 & \textbf{0.1209} & \textbf{SheafStain (ours)} & \textbf{0.766} & \textbf{0.845} & \textbf{0.794} \\
    \bottomrule
  \end{tabular}%
  }
\end{table}

\begin{table}[t]
  \caption{\textbf{Quantitative comparison on BCI and MIST}. Best metrics in bold, second-best metrics underlined.}
  \label{tab:comparison}
  \centering
  \footnotesize
  \setlength{\tabcolsep}{3pt}
  \resizebox{\textwidth}{!}{%
  \begin{tabular}{lllcccccc}
    \toprule
    Dataset & Stain & Method & FID$\downarrow$ & KID$\times10^{3}\downarrow$ & LPIPS$\downarrow$ & DISTS$\downarrow$ & TS$\downarrow$ & DAB-$r$$\uparrow$ \\
    \midrule
    \multirow{7}{*}{BCI} & \multirow{7}{*}{HER2}
     & pix2pix                                  & 172.7932 & 138.9050 & 0.4800{\scriptsize$\pm$0.0722} & 0.2753{\scriptsize$\pm$0.0698} & 0.0533{\scriptsize$\pm$0.0214} & 0.0559 \\
     & & CycleGAN                               & 96.3496 & 50.8020 & 0.5448{\scriptsize$\pm$0.0659} & 0.2920{\scriptsize$\pm$0.0599} & 0.0475{\scriptsize$\pm$0.0262} & 0.0668 \\
     & & PSPStain                               & 185.6991 & 173.7360 & 0.5695{\scriptsize$\pm$0.0640} & 0.2924{\scriptsize$\pm$0.0494} & 0.0885{\scriptsize$\pm$0.0354} & 0.0317 \\
     & & D-VST                                  & 87.3100 & 62.7670 & 0.5107{\scriptsize$\pm$0.0707} & 0.2570{\scriptsize$\pm$0.0596} & \underline{0.0164{\scriptsize$\pm$0.0070}} & 0.0736 \\
     & & UNIStainNet                            & \underline{67.6322} & \underline{29.3000} & \textbf{0.4577{\scriptsize$\pm$0.0711}} & \underline{0.2279{\scriptsize$\pm$0.0531}} & 0.0448{\scriptsize$\pm$0.0320} & \textbf{0.1267} \\
     & & UNSB                                   & 227.7889 & 236.2310 & 0.6635{\scriptsize$\pm$0.0677} & 0.3347{\scriptsize$\pm$0.0452} & 0.1464{\scriptsize$\pm$0.0161} & 0.0589 \\
     & & \textbf{SheafStain (ours)} & \textbf{36.3626} & \textbf{4.2070} & \underline{0.4689{\scriptsize$\pm$0.0584}} & \textbf{0.2132{\scriptsize$\pm$0.0479}} & \textbf{0.0146{\scriptsize$\pm$0.0072}} & \underline{0.1209} \\
    \midrule
    \multirow{28}{*}{MIST} & \multirow{7}{*}{HER2}
     & pix2pix              & 186.3988 & 186.3800 & 0.5631{\scriptsize$\pm$0.0619} & 0.3624{\scriptsize$\pm$0.0685} & 0.0851{\scriptsize$\pm$0.0196} & \underline{0.0866} \\
     & & CycleGAN             & 95.9094 & 54.0670 & 0.5766{\scriptsize$\pm$0.0548} & 0.2791{\scriptsize$\pm$0.0454} & 0.0698{\scriptsize$\pm$0.0110} & 0.0681 \\
     & & PSPStain             & \underline{50.1078} & \underline{10.6750} & \underline{0.5371{\scriptsize$\pm$0.0594}} & \underline{0.2466{\scriptsize$\pm$0.0388}} & 0.0763{\scriptsize$\pm$0.0210} & 0.0841 \\
     & & D-VST                & 94.0706 & 65.4310 & 0.5578{\scriptsize$\pm$0.0525} & 0.2613{\scriptsize$\pm$0.0462} & \underline{0.0336{\scriptsize$\pm$0.0062}} & 0.0356 \\
     & & UNIStainNet          & 101.7560 & 74.9900 & 0.5978{\scriptsize$\pm$0.0617} & 0.3037{\scriptsize$\pm$0.0796} & 0.1177{\scriptsize$\pm$0.0257} & 0.0606 \\
     & & UNSB                 & 57.9643 & 20.5260 & 0.5692{\scriptsize$\pm$0.0640} & 0.2808{\scriptsize$\pm$0.0547} & 0.1002{\scriptsize$\pm$0.0307} & 0.0310 \\
     & & \textbf{SheafStain (ours)} & \textbf{34.5080} & \textbf{2.2350} & \textbf{0.5196{\scriptsize$\pm$0.0646}} & \textbf{0.2056{\scriptsize$\pm$0.0382}} & \textbf{0.0292{\scriptsize$\pm$0.0080}} & \textbf{0.1187} \\
    \cmidrule(lr){2-9}
     & \multirow{7}{*}{ER}
     & pix2pix              & 208.5043 & 210.1640 & 0.5735{\scriptsize$\pm$0.0657} & 0.4237{\scriptsize$\pm$0.0815} & 0.0692{\scriptsize$\pm$0.0259} & 0.0551 \\
     & & CycleGAN             & 67.1467 & 28.0530 & 0.5492{\scriptsize$\pm$0.0549} & 0.2735{\scriptsize$\pm$0.0616} & 0.0693{\scriptsize$\pm$0.0228} & 0.0884 \\
     & & PSPStain             & \underline{45.9200} & 16.0910 & \underline{0.5235{\scriptsize$\pm$0.0496}} & 0.2534{\scriptsize$\pm$0.0493} & 0.0752{\scriptsize$\pm$0.0286} & \underline{0.0906} \\
     & & D-VST                & 84.2247 & 65.3180 & 0.5518{\scriptsize$\pm$0.0384} & 0.2598{\scriptsize$\pm$0.0632} & \underline{0.0310{\scriptsize$\pm$0.0063}} & 0.0554 \\
     & & UNIStainNet          & 74.8386 & 37.2410 & 0.5259{\scriptsize$\pm$0.0550} & \underline{0.2461{\scriptsize$\pm$0.0541}} & 0.0571{\scriptsize$\pm$0.0112} & 0.0497 \\
     & & UNSB                 & 49.2278 & \underline{16.0520} & 0.5659{\scriptsize$\pm$0.0545} & 0.2710{\scriptsize$\pm$0.0606} & 0.0924{\scriptsize$\pm$0.0279} & 0.0790 \\
     & & \textbf{SheafStain (ours)} & \textbf{29.0824} & \textbf{2.5000} & \textbf{0.5042{\scriptsize$\pm$0.0619}} & \textbf{0.1934{\scriptsize$\pm$0.0358}} & \textbf{0.0292{\scriptsize$\pm$0.0075}} & \textbf{0.1248} \\
    \cmidrule(lr){2-9}
     & \multirow{7}{*}{PR}
     & pix2pix              & 210.9157 & 206.3250 & 0.5788{\scriptsize$\pm$0.0639} & 0.4148{\scriptsize$\pm$0.0754} & 0.0686{\scriptsize$\pm$0.0331} & 0.0656 \\
     & & CycleGAN             & 140.6118 & 99.8560 & 0.5783{\scriptsize$\pm$0.0474} & 0.3046{\scriptsize$\pm$0.0621} & 0.1022{\scriptsize$\pm$0.0238} & 0.0823 \\
     & & PSPStain             & \underline{48.6565} & \underline{19.2100} & 0.5447{\scriptsize$\pm$0.0463} & 0.2744{\scriptsize$\pm$0.0564} & 0.0797{\scriptsize$\pm$0.0287} & \underline{0.0905} \\
     & & D-VST                & 90.1997 & 67.5810 & 0.5566{\scriptsize$\pm$0.0400} & \underline{0.2650{\scriptsize$\pm$0.0631}} & \underline{0.0308{\scriptsize$\pm$0.0064}} & 0.0502 \\
     & & UNIStainNet          & 91.2102 & 60.3800 & \underline{0.5232{\scriptsize$\pm$0.0729}} & 0.3060{\scriptsize$\pm$0.0916} & 0.0351{\scriptsize$\pm$0.0059} & 0.0429 \\
     & & UNSB                 & 54.2625 & 21.3240 & 0.5741{\scriptsize$\pm$0.0594} & 0.2797{\scriptsize$\pm$0.0639} & 0.0904{\scriptsize$\pm$0.0368} & 0.0544 \\
     & & \textbf{SheafStain (ours)} & \textbf{29.7240} & \textbf{2.1910} & \textbf{0.5017{\scriptsize$\pm$0.0588}} & \textbf{0.1962{\scriptsize$\pm$0.0361}} & \textbf{0.0272{\scriptsize$\pm$0.0075}} & \textbf{0.1184} \\
    \cmidrule(lr){2-9}
     & \multirow{7}{*}{Ki-67}
     & pix2pix              & 253.8902 & 306.2140 & 0.5707{\scriptsize$\pm$0.0371} & 0.3617{\scriptsize$\pm$0.0422} & 0.0839{\scriptsize$\pm$0.0180} & 0.0257 \\
     & & CycleGAN             & 74.4676 & 43.0280 & 0.5552{\scriptsize$\pm$0.0433} & 0.2575{\scriptsize$\pm$0.0408} & 0.0618{\scriptsize$\pm$0.0164} & \underline{0.0423} \\
     & & PSPStain             & 44.0905 & 15.9350 & 0.5442{\scriptsize$\pm$0.0345} & 0.2618{\scriptsize$\pm$0.0383} & 0.0710{\scriptsize$\pm$0.0226} & 0.0340 \\
     & & D-VST                & 74.1956 & 54.2740 & 0.5587{\scriptsize$\pm$0.0324} & 0.2414{\scriptsize$\pm$0.0376} & \underline{0.0282{\scriptsize$\pm$0.0068}} & 0.0235 \\
     & & UNIStainNet          & 62.7007 & 37.8290 & \underline{0.5420{\scriptsize$\pm$0.0359}} & \underline{0.2254{\scriptsize$\pm$0.0361}} & 0.0620{\scriptsize$\pm$0.0209} & 0.0200 \\
     & & UNSB                 & \underline{38.1956} & \underline{10.1780} & 0.5539{\scriptsize$\pm$0.0351} & 0.2514{\scriptsize$\pm$0.0397} & 0.0857{\scriptsize$\pm$0.0225} & 0.0326 \\
     & & \textbf{SheafStain (ours)} & \textbf{25.3978} & \textbf{3.9220} & \textbf{0.5269{\scriptsize$\pm$0.0404}} & \textbf{0.2124{\scriptsize$\pm$0.0324}} & \textbf{0.0261{\scriptsize$\pm$0.0066}} & \textbf{0.0493} \\
    \bottomrule
  \end{tabular}%
  }
\end{table}

We compare SheafStain against six prior methods on BCI and MIST datasets, and
Table~\ref{tab:comparison} reports performance at the $1024 \times 1024$ aggregated image level. SheafStain attains the best FID, KID, DISTS, and TS
on every cell, the best DAB-$r$ on the four MIST cells, and the best LPIPS on every cell except BCI HER2
(UNIStainNet leads by $0.011$). Distribution-metric margins are multiplicative.
KID$\times 10^{3}$ is $2.6$--$8.8\times$ below the strongest prior method and FID is
$30$--$46\%$ below the runner-up. On stitching, D-VST is the closest competitor,
yet SheafStain leads by $6$--$13\%$ across blocks.
The DAB-$r$ margin on the four MIST cells is $1.2$--$1.4\times$ over the runner-up, while on BCI HER2
UNIStainNet leads at $0.1267$ with SheafStain following at $0.1209$.
Additional metric set results are in Appendix~\ref{app:full_comparison}.

\begin{figure}[t]
  \centering
  \includegraphics[width=\linewidth]{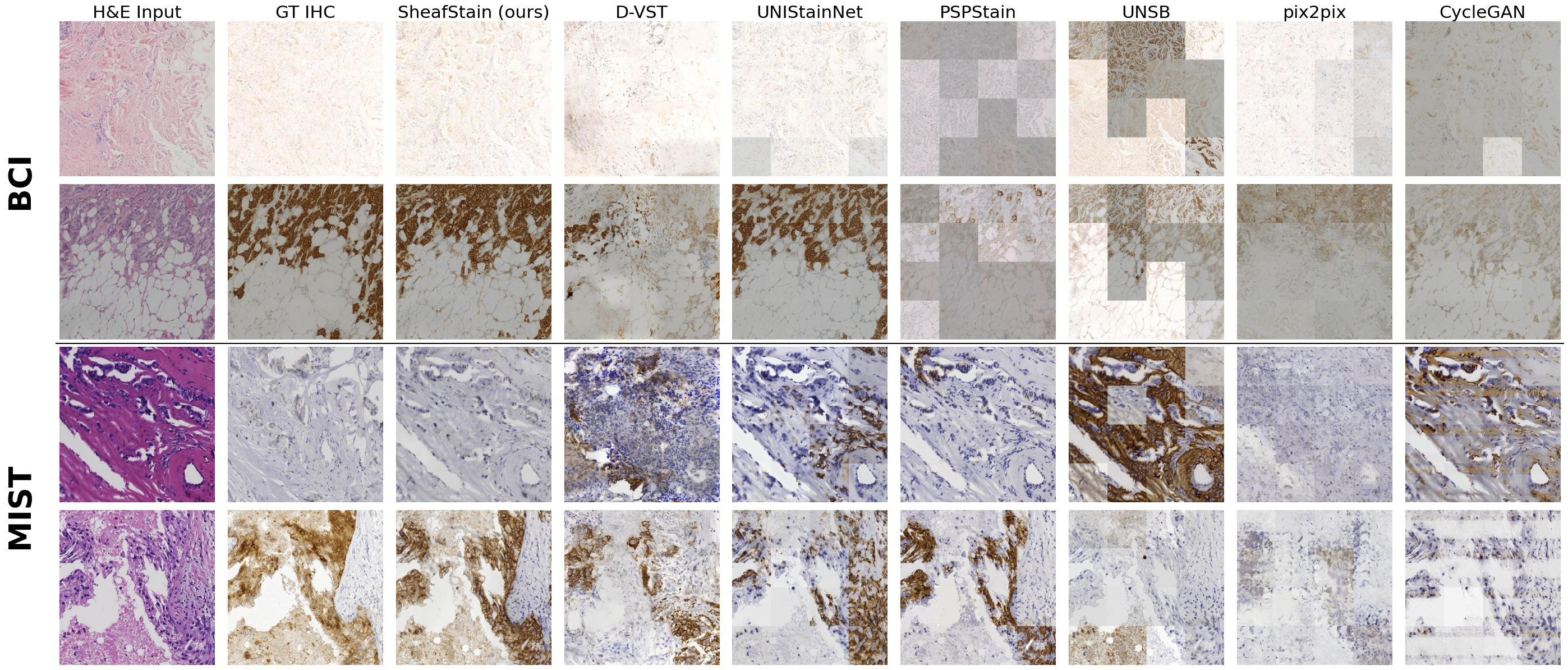}
  \caption{\textbf{Qualitative comparison on BCI and MIST HER2 datasets.}
           The first MIST row uses the pairing the dataset itself provides. The tissue
           structure agrees between the two images while the color does not, because that field
           carries little chromogen, so its IHC reference is dominated by counterstain while the
           H\&E is strongly eosinophilic. MIST pairs are also cut from consecutive sections, so
           every reference is a neighboring slice rather than the same physical plane.}
  \label{fig:qualitative}
\end{figure}

Figure~\ref{fig:qualitative} shows
that SheafStain produces the least visible patch-boundary seams and best matches GT IHC's DAB
chromogen intensity and spatial distribution, with additional results in
Appendix~\ref{app:qualitative}. Exact pixel-level agreement with the GT is not attainable by any
method, since it comes from a different physical section, and
Appendix~\ref{app:full_comparison} quantifies the resulting bound. We, therefore, test the outputs
with the downstream protocol above, which does not depend on appearance.

\section{Conclusion}
\label{sec:conclusion}
We present SheafStain, a sheaf-theoretic framework for patch-wise virtual staining
that unifies the Schr\"{o}dinger Bridge with local-to-global consistency in VFM embedding space. 
We identify \emph{context contamination} as the structural reason VFM embeddings fail the
overlap compatibility that gluing requires. SheafStain addresses this without an
external correction module: VFM embeddings are interpreted as sheaf sections, tissue
context is injected via a spatial conditioning map and neighborhood CLS token, and
pixel sheaf and \v{C}ech-cocycle losses enforce local-to-global coherence, so the generator 
learns \emph{inherent restriction-map} behavior. On BCI and MIST datasets 
across HER2, ER, PR, and Ki-67, SheafStain outperforms six prior methods on the
majority of metrics while mitigating patch-boundary stitching artifacts.

\paragraph{Limitations.}
We note four limitations of SheafStain.
First, validation is restricted to BCI and MIST. External validation on real-world clinical datasets
remains future work. Second, our evaluation is conducted at the conventional $1024 \times 1024$ scale,
sufficient to expose the inter-patch consistency but smaller than WSI.
Third, although the sheaf-theoretic formulation is organ-agnostic, it is not empirically verified
on other organs (e.g., colorectal, lung, and prostate).
Fourth, inference is computationally expensive. The VFM forward accounts for about $90\%$ of the
per-image cost, which raises it well above the faster prior methods, although
Appendix~\ref{app:compute_cost} places \method{} at the high-quality endpoint of the
Pareto frontier. The VFM is a swappable component, so the total cost depends on which
encoder is used.

\paragraph{Future works.}
The first three limitations motivate three planned extensions. Prior virtual staining studies similarly
stop short of WSI quantitative evaluation. We plan comprehensive WSI-level comparisons as the
next step, which will also show whether the framework carries to other patch-assembled
whole-slide tasks. Scaling the evaluation also calls for further public cohorts. IHC4BC~\citep{ihc4bc} is public and
covers the same four markers as MIST, which makes it the next dataset to add. Adding it means retraining all seven methods on about $90{,}000$ pairs from
$240$ slide pairs. HER2match~\citep{her2match} is a further candidate, though its same-section
re-staining removes the weak pairing the method is built to handle.
We further plan cross-organ generalization on additional tissues and prospective
clinical-utility assessment (e.g., HER2 scoring concordance under sheaf-consistent outputs).

\begin{ack}
This work was supported by the Technology Innovation Program
(RS-2025-02221011, Development of Medical-Specialized Multimodal Hyperscale Generative AI Technology for Global Integration)
funded by the Ministry of Trade Industry \& Energy (MOTIE, South Korea).

Hyeongyeol Lim, Hongjun Yoon, Eunjin Jang, Daeky Jeong, and Won~June~Cho are employees of DEEPNOID Inc.
The authors declare no other competing interests.
\end{ack}

\bibliographystyle{unsrtnat}

\clearpage

\appendix

\section*{Appendix}
\begingroup
\hypersetup{pdfborder={0 0 0}}
\newcommand{\apptocsec}[1]{%
  \par\vspace{16pt}\noindent\hyperref[#1]{\makebox[1.6em][l]{\textbf{\ref*{#1}}}\textbf{\nameref*{#1}}}%
  \hfill\textbf{\pageref*{#1}}\par}
\newcommand{\apptocsub}[1]{%
  \par\vspace{7pt}\noindent\hspace*{1.6em}\hyperref[#1]{\makebox[2.4em][l]{\ref*{#1}}\nameref*{#1}}%
  \dotfill\pageref*{#1}\par}
\apptocsec{app:sheaf}
\apptocsec{app:theory}
\apptocsec{app:loss}
\apptocsec{app:cs-vfm-align}
\apptocsec{app:training}
\apptocsec{app:implementation}
\apptocsec{app:inference}
\apptocsec{app:datasets}
\apptocsec{app:gluing}
\apptocsec{app:full_comparison}
\apptocsec{app:qualitative}
\apptocsec{app:downstream}
\apptocsub{appx:downstream:protocol}
\apptocsub{appx:downstream:roc}
\apptocsec{app:compute_cost}
\apptocsub{appx:compcost:protocol}
\apptocsub{appx:compcost:pareto}
\apptocsub{appx:compcost:vfm}
\apptocsub{appx:compcost:caveats}
\apptocsub{appx:compcost:discussion}
\endgroup
\clearpage


\section{Sheaf Theory Background}
\label{app:sheaf}

This section expands upon the brief sheaf-theoretic background provided 
in Section~\ref{sec:prelim}. Following standard treatments~\citep{bredon1997sheaf,sheaf_survey}, 
we review only the specific constructions utilized in this work.

\paragraph{Presheaf, sheaf axioms, and restriction maps.}
Let $X$ be a topological space. A \emph{presheaf} $\sheafF$ on $X$
assigns to each open set $U \subseteq X$ a vector space $\sheafF(U)$.
An element $s \in \sheafF(U)$ is called a \emph{section} of $\sheafF$
over $U$. The elements of $\sheafF(U)$ are related across inclusions
by linear maps that we formalize below as restriction maps. A
presheaf is further a \emph{sheaf} if for every open $V \subseteq X$ and every open cover $\mathcal{U} = \{U_i\}_{i \in I}$ of $V$, the following hold:
\begin{enumerate}[leftmargin=*,itemsep=2pt]
  \item \textbf{Locality:} If $s, t \in \sheafF(V)$ have identical
    restrictions to every $U_i$ ($i \in I$), then $s = t$.
  \item \textbf{Gluing:} If a family of local sections
    $\{s_i \in \sheafF(U_i)\}$ is pairwise compatible on overlaps,
    there exists a unique $s \in \sheafF(V)$ whose restriction to
    each $U_i$ equals $s_i$.
\end{enumerate}
Locality says a section is determined by its local
restrictions. Gluing says compatible local sections assemble into a
unique section. The compatibility condition and the
notion of ``restriction to $U_i$'' are formalized by the \emph{restriction
map} $\rho_{U \to V}$, defined for any inclusion
$V \subseteq U$ as
\begin{equation}
  \rho_{U \to V}: \sheafF(U) \to \sheafF(V),
  \label{eq:appendix_restriction}
\end{equation}
satisfying compositionality
$\rho_{U \to W} = \rho_{V \to W} \circ \rho_{U \to V}$ whenever
$W \subseteq V \subseteq U$, with shorthand $s|_V := \rho_{U \to V}(s)$.
Local sections $\{s_i\}$ on a cover $\mathcal{U}$ are
\emph{compatible} when restrictions agree on every pairwise overlap:
\begin{equation}
  \rho_{U_i \to U_{ij}}(s_i) = \rho_{U_j \to U_{ij}}(s_j),
  \quad \forall\, i, j \text{ with } U_{ij} \neq \emptyset,
  \label{eq:compatibility}
\end{equation}
where $U_{ij} := U_i \cap U_j$. Eq.~\ref{eq:compatibility} is the
compatibility condition SheafStain enforces in pixel space via
Eq.~\ref{eq:sheaf_loss}. When sections fail this compatibility,
the deviation is measured by the \emph{sheaf Laplacian}
$\Delta_0 = \delta^*\delta$, where $\delta$ is the coboundary map
encoding restriction maps~\citep{nsd, sheaf_survey}. The associated
\emph{Dirichlet energy}
$E(x) = \langle \Delta_0 x, x \rangle \ge 0$ quantifies the total
inconsistency across all overlaps, and driving $E(x) \to 0$ enforces compatibility on every overlap, so that $x$ behaves as a globally consistent section.

The formal stalk is $\sheafF_x := \varinjlim_{x \in U} \sheafF(U)$, the direct limit at a point, 
whose elements are germs. In this work, a patch embedding $s_i \in \sheafF(U_i)$ is formally a section over $U_i$, 
while an individual per-position representation of the spatial conditioning map (Section~\ref{sec:conditioning}) is a value in the corresponding fiber. 
All statements involving restriction maps are phrased in terms of sections $s_i$.

\paragraph{\v{C}ech cohomology and the SheafStain cocycle reduction.}
For an open cover $\mathcal{U} = \{U_i\}_{i \in I}$ and a presheaf
$\sheafF$, \v{C}ech cohomology assembles pairwise discrepancies on
overlaps into a global obstruction class. The zeroth group
$\check{H}^0(\mathcal{U}, \sheafF)$ collects the section families that are
compatible on all overlaps.
Concretely, a \emph{0-cochain} on $\mathcal{U}$ is a family
$\{h_i \in \sheafF(U_i)\}$, and its coboundary
$\delta\{h_i\}_{ij} := h_i|_{U_{ij}} - h_j|_{U_{ij}}$ records pairwise
disagreement on overlaps. A 0-cochain is a \emph{0-cocycle} when this
coboundary vanishes, i.e., $h_i|_{U_{ij}} = h_j|_{U_{ij}}$ for every
$i, j$, the compatibility condition (Eq.~\ref{eq:compatibility}).
Thus $\check{H}^0$ is the space of 0-cocycles, that is, of section families
that agree on every overlap. When $\sheafF$ is a sheaf these are exactly the
global sections, whereas for a presheaf they need not be.
The first group $\check{H}^1(\mathcal{U}, \sheafF)$ measures the
\emph{obstruction} to gluing. It is the quotient of 1-cochains
$\{g_{ij} \in \sheafF(U_{ij})\}$ satisfying the \emph{1-cocycle
condition}
\begin{equation}
  g_{ij} + g_{jk} = g_{ik} \quad \text{on } U_{ijk} := U_i \cap U_j \cap U_k,
  \label{eq:cocycle_condition}
\end{equation}
modulo \emph{coboundaries} of the form
$g_{ij} = h_i|_{U_{ij}} - h_j|_{U_{ij}}$ arising from a 0-cochain
$\{h_i \in \sheafF(U_i)\}$. Vanishing of $\check{H}^1(\mathcal{U}, \sheafF)$ means every
1-cocycle on this particular cover is a coboundary, so the pairwise discrepancies
can be absorbed by adjusting the local sections. A non-trivial class is an
obstruction that no such adjustment removes.

We apply this framework to SheafStain outputs. Let
$G_i := G_\theta(P_i, t, M_i, c)$ denote the generator output on
patch $P_i$. The pairwise discrepancy on the overlap
$\mathcal{O} = U_i \cap U_j$ is
$g_{ij} := G_i|_\mathcal{O} - G_j|_\mathcal{O}$, and the pixel sheaf
loss (Eq.~\ref{eq:sheaf_loss}) directly penalizes $\|g_{ij}\|_1$
together with a tone-matching term. On the triple overlap
$\mathcal{O}_3 = U_{\text{ref}} \cap U_{\text{adj}_1} \cap U_{\text{adj}_2}$,
the three pairwise discrepancies $g_{12}$, $g_{13}$, and $g_{23}$ are not
independent. They satisfy the cocycle relation
$g_{12} + g_{23} = g_{13}$ \emph{trivially}, as $\{g_{ij}\}$ is a coboundary by construction (telescoping
cancellation $(G_1 - G_2) + (G_2 - G_3) = G_1 - G_3$). The pixel
sheaf loss and the cocycle loss (Eq.~\ref{eq:cocycle_loss}) penalize
$g_{12}$ and the remaining two terms $g_{13}$ and $g_{23}$, respectively.
Together they drive all three
discrepancies toward zero, so that the sections $\{G_i\}$ approximately form a 0-cocycle on the cover 
(i.e., a section family approximately compatible on all overlaps). 
The cocycle class $[\{g_{ij}\}] \in \check{H}^1$ is trivially zero for the same reason. 
This enforces the necessary condition for gluing cited in Section~\ref{sec:sheaf_loss}. 
The residual deviation from compatibility is controlled by the magnitude of the learned discrepancies $\|g_{ij}\|_1$, 
so the generator's outputs approach a compatible family on the open cover.

\section{SheafStain as a Regularized Schr\"odinger Bridge}
\label{app:theory}

This section expands the one-sentence positioning of
Section~\ref{sec:prelim} into a self-contained account of how the
sheaf and cocycle regularizers (Section~\ref{sec:sheaf_loss}) sit
within the entropy-regularized optimal transport (OT) family, and why generator-side
regularization suffices.

\paragraph{Schr\"odinger Bridge and UNSB.}
OT seeks the most efficient way to transform one
probability distribution into another, formalized as minimizing a
transport cost over the set $\Pi(\pi_0, \pi_1)$ of couplings (joint
distributions with marginals $\pi_0, \pi_1$)~\citep{villani2009ot}.
The SB problem~\citep{schrodinger1932}
introduces an entropic regularization to OT. Given reference dynamics
(typically a Brownian motion with diffusion coefficient $\tau$), SB
finds the stochastic process $Q$ that transports $\pi_0$ to $\pi_1$
while remaining closest to the reference process in KL
divergence~\citep{leonard2013sb}:
\begin{equation}
  Q_{\mathrm{SB}} = \arg\min_{Q \in \mathcal{D}(\pi_0, \pi_1)}
                    \mathrm{KL}(Q \,\|\, W^\tau),
  \label{eq:sb_objective}
\end{equation}
where $\mathcal{D}(\pi_0, \pi_1)$ is the set of stochastic processes
with marginals $\pi_0$ at $t = 0$ and $\pi_1$ at $t = 1$, and $W^\tau$
denotes the Wiener process with diffusion coefficient $\tau$. As
$\tau \to 0$, the SB solution converges to the deterministic OT
map~\citep{chen2021sbotconnection}, so $\tau$ controls the
trade-off between transport optimality and stochastic diffusion.
Classical SB algorithms (e.g., Sinkhorn-type iterations and their
continuous analogs~\citep{chen2021sbotconnection}) scale poorly to
high-dimensional image spaces.

UNSB~\citep{unsb} addresses this via the \emph{self-similarity} property of
SB (the restriction of an SB to any sub-interval $[t_i, 1]$ is
itself an SB), decomposing the problem into a sequence of
adversarial sub-problems indexed by time $t_i$. At each $t_i$, a
generator $q_\phi$ predicts the forward SDE transition under the
entropy-regularized SB objective
\begin{equation}
  \mathcal{L}_{\text{SB}}(\phi, t_i)
   = \mathbb{E}\!\left[\|x_{t_i} - x_1\|^2\right]
     - 2\tau(1 - t_i)\, H\!\left(q_\phi\right),
  \label{eq:unsb_sbloss}
\end{equation}
combining a squared transport cost with an entropy term of weight
$\tau$. Marginal matching is supplied by a discriminator $D$ via a
Kantorovich-dual divergence estimate~\citep{unsb}, and the entropy
$H(q_\phi)$ is approximated by an auxiliary energy network $E$ trained
with a contrastive-divergence objective. The combined adversarial
scheme eliminates the need for paired training data, making UNSB
well-suited for unpaired image-to-image translation, including virtual
staining where H\&E and IHC images are obtained from consecutive
tissue sections without pixel-level correspondence.

\paragraph{SheafStain as a regularized-SB problem.}
SheafStain augments the SB objective (Eq.~\ref{eq:sb_objective}) with
additional convex regularizers on the coupling plan, the pixel sheaf
loss (Eq.~\ref{eq:sheaf_loss}) and the cocycle loss
(Eq.~\ref{eq:cocycle_loss}). Schematically, the SheafStain coupling solves
\begin{equation}
  Q^{\ast} \in \arg\min_{Q \in \mathcal{D}(\pi_0, \pi_1)}
    \Big[\, \mathrm{KL}(Q \,\|\, W^\tau)
       + \sum_{k} \lambda_k\, \mathcal{R}_k(Q) \,\Big],
  \label{eq:reg_sb}
\end{equation}
where $\{\mathcal{R}_k\}$ ranges over the sheaf-derived structural
regularizers (and the optional pathology-specific terms in
Appendix~\ref{app:loss}), with weights $\lambda_k$ matching those of
the generator objective in Eq.~\ref{eq:total_loss}.

\paragraph{Marginal preservation.}
A key consistency property is that $\{\mathcal{R}_k\}$ act on the
structure of the coupling plan but leave the marginal-constraint set
$\mathcal{D}(\pi_0, \pi_1)$ unchanged. The sheaf and cocycle losses
depend on patch-pair outputs $G_i, G_j$ without altering the
source/target marginals. Eq.~\ref{eq:reg_sb} therefore remains a
valid SB variant within the entropy-regularized OT family, and any
feasible solution still couples the H\&E and IHC distributions
correctly. Equivalently, from the OT perspective,
$\mathcal{L}_{\text{sheaf}}$ and $\mathcal{L}_{\text{cocycle}}$ act
as structural constraints on the transport plan. They restrict the
coupling to one that respects patch-level sheaf consistency while
preserving the marginal-matching structure enforced by the SB
objective. Because the discriminator $D$ and energy network $E$
inherited from the baseline are co-trained adversarially with $G_\theta$,
they adapt to the SheafStain-modified primal without explicit redesign.
Under the combined objective, the raw SB-loss exhibits the
standard min-max oscillation, so sample-based metrics (FID, KID, TS)
provide the appropriate convergence certificate.

\section{Pathology-Specific Regularizers}
\label{app:loss}

To complement the sheaf-consistency objectives we utilize two pathology-specific regularizers,
$\mathcal{L}_{\text{DAB}}$ and $\mathcal{L}_{\text{fourier}}$, both introduced in
Section~\ref{sec:ablation}. The output-side VFM alignment
($\mathcal{L}_{\text{stain-align}}$) is described separately in Appendix~\ref{app:cs-vfm-align}.
$\mathcal{L}_{\text{DAB}}$ matches the mean top-10\% intensity
of the deconvolved DAB channel between $G_{\text{ref}}$ and the target IHC patch,
providing a chromogen-fidelity signal that is translation-invariant under weak pairing.
$\mathcal{L}_{\text{fourier}}$ penalizes discrepancies in the high-frequency
log-magnitude spectrum between grayscale conversions of $G_{\text{ref}}$
and the target, preserving glandular boundaries and stromal texture.

Both operate on translation-invariant signals and are robust
to spatial misalignment under weakly-paired H\&E--IHC consecutive sections.

\paragraph{DAB intensity loss.}
IHC images use diaminobenzidine (DAB) as the primary chromogen for biomarker visualization. We extract the DAB channel via Beer--Lambert color deconvolution~\citep{ruifrok2001} and match the mean top-10\% DAB intensity (p90) between generated and target images:
\begin{equation}
  \mathcal{L}_{\text{DAB}} = \left| \mathrm{p90}\big(\mathrm{DAB}(G_{\text{ref}})\big) - \mathrm{p90}\big(\mathrm{DAB}(y)\big) \right|
  \label{eq:dab_loss}
\end{equation}
where $\mathrm{DAB}(\cdot)$ denotes DAB-channel extraction and $\mathrm{p90}(\cdot)$ is the mean intensity over pixels above the 90th percentile. The target score is detached from the computation graph, so gradients flow only to the generator. The top-percentile design is translation-invariant. It captures global staining intensity regardless of spatial position, making it robust to misalignment between weakly-paired H\&E and IHC sections.

\paragraph{Fourier edge loss.}
Virtual staining must preserve fine histological structures (e.g., glandular boundaries, nuclear contours, and stromal texture) that reside in the high-frequency spectrum. We adopt a Fourier-domain loss that penalizes high-frequency log-magnitude discrepancy between generated and target images:
\begin{equation}
  \mathcal{L}_{\text{fourier}} = \left\| H \odot \log\!\big(1 + |\mathrm{FFT}(G_{\text{gray}})|\big) - H \odot \log\!\big(1 + |\mathrm{FFT}(y_{\text{gray}})|\big) \right\|_1
  \label{eq:fourier_loss}
\end{equation}
where $G_{\text{gray}}, y_{\text{gray}}$ are grayscale conversions, $\mathrm{FFT}$ denotes the 2D Fourier transform, and $H$ is a radial high-pass mask retaining frequencies whose radial distance from the spectrum center exceeds $25\%$ of the maximum radius. The $\log(1+\cdot)$ scaling ensures numerical stability and perceptually meaningful weighting. Since the Fourier magnitude spectrum is translation-invariant ($|\mathrm{FFT}(\mathrm{shift}(x))| = |\mathrm{FFT}(x)|$), this loss is inherently robust to spatial misalignment. The functional form is adapted from UNIStainNet's edge-preservation regularizer~\citep{unistainnet}.

\section{Vision Foundation Model Alignment}
\label{app:cs-vfm-align}

VFM Alignment is the output-side counterpart of SheafStain's
VFM conditioning. The same VFM that supplies input-side
conditioning (spatial map $M$ and neighborhood CLS token $c_{n}$ in
Section~\ref{sec:method}) also serves as a feature-space supervisor at the
output, ensuring section consistency is enforced at \emph{both} ends of the
H\&E$\to$IHC translation. The procedure complements the pixel-level DAB
intensity loss ($\mathcal{L}_{\text{DAB}}$, Appendix~\ref{app:loss}) by
aligning the spatial layout of chromogen, not just its scalar statistic.

Concretely, we (i) extract the DAB channel from both generated and target
images via Beer--Lambert color deconvolution~\citep{ruifrok2001},
(ii) re-render each as a brown-tinted RGB image $\mathrm{rgb}_{\text{DAB}}(\cdot)$
by applying the inverse stain transform to the DAB-only intensity
(chromogen only, on a white background), and (iii) match their VFM spatial features:
\begin{equation}
  \mathcal{L}_{\text{stain-align}}
  = \big\| \phi_{\text{VFM}}\!\big(\mathrm{rgb}_{\text{DAB}}(G_{\text{ref}})\big)
  - \phi_{\text{VFM}}\!\big(\mathrm{rgb}_{\text{DAB}}(y)\big) \big\|_1 ,
  \label{eq:stain_align_loss}
\end{equation}
where $\phi_{\text{VFM}}$ denotes per-token spatial features from the frozen
Prov-GigaPath ViT and the target branch is detached.
From the sheaf-theoretic perspective, this closes the consistency loop by
aligning post-translation sections within the same VFM feature space that
defines the input sections. Empirically (Section~\ref{sec:ablation}), VFM
Alignment gives a small but consistent gain on DAB-$r$
($+2.2\%$ on BCI/HER2) while also lowering FID, LPIPS,
and DISTS.


\section{Training Procedure}
\label{app:training}

This section gives the full SheafStain generator objective, the inherent
restriction-map-learning argument, and the per-step training procedure
presented in Section~\ref{sec:method} and Figure~\ref{fig:flow}.

\paragraph{Total objective.}
The full SheafStain generator objective combines the baseline's standard losses with our sheaf and pathology-specific contributions:
\begin{equation}
  \mathcal{L}_{G} = \mathcal{L}_{\text{GAN}} + \sum_{k \in \mathcal{K}} \lambda_k \mathcal{L}_k,
  \quad \mathcal{K} = \{\text{SB},\ \text{NCE},\ \text{sheaf},\ \text{cocycle},\ \text{fourier},\ \text{DAB},\ \text{stain-align}\},
  \label{eq:total_loss}
\end{equation}
where $\mathcal{L}_{\text{GAN}}$ (adversarial), $\mathcal{L}_{\text{SB}}$,
and $\mathcal{L}_{\text{NCE}}$ (Noise Contrastive Estimation) are inherited from the baseline
and the remaining terms are our contributions.

\paragraph{Loss weights and curriculum.}
We use $\lambda_{\text{SB},\,\text{NCE},\,\text{sheaf}} = 1$,
$\lambda_{\text{cocycle},\,\text{DAB}} = 0.1$, and $\lambda_{\text{fourier}} = 0.5$. The pixel sheaf balance coefficient is $\alpha = 1$
(Eq.~\ref{eq:sheaf_loss}).
Training runs for 400 epochs total (200 fixed + 200 linear decay).
VFM Alignment ($\mathcal{L}_{\text{stain-align}}$, Appendix~\ref{app:cs-vfm-align})
is activated only for the final 50 epochs. We set $\lambda_{\text{stain-align}} = 0$ for
epochs 1 to 350 and $\lambda_{\text{stain-align}} = 1$ for epochs 351 to 400.
All other weights are fixed throughout.

\paragraph{Hyperparameter selection.}
The seven weights of Eq.~\ref{eq:total_loss} were not all set in the same way.
$\lambda_{\text{SB}}$, $\lambda_{\text{NCE}}$, and $\lambda_{\text{sheaf}}$ are fixed at $1$ and
were not varied in any run. UNSB sets its bridge weight to $1$, and the Baseline (UNSB) row of
Table~\ref{tab:ablation} is trained with the same weights we use, so the ablation adds our
components without changing the baseline weighting. $\lambda_{\text{sheaf}}$ is fixed at $1$ by
construction, since the sheaf term is the objective this work is about and therefore carries the
unit scale against which the auxiliary weights are expressed. $\lambda_{\text{stain-align}}$ is
$1$ whenever the term is active, so what we set for it is a schedule rather than a magnitude, as
described above.

The remaining weights were set in two stages. The first stage is the component ablation of
Table~\ref{tab:ablation}, which decides which terms belong in the objective at all. The second
stage uses a coarse grid search over $\alpha$, $\lambda_{\text{cocycle}}$,
and $\lambda_{\text{fourier}}$ in $\{0.1, 0.5, 1.0\}$ with $\lambda_{\text{DAB}}$ held at $0.1$,
giving $3^3 = 27$ settings on BCI (Table~\ref{tab:weight_grid}).

\begin{table}[h]
  \centering
  \caption{\textbf{Weight grid over the auxiliary terms.} All $27$ settings of
  $\alpha$, $\lambda_{\text{cocycle}}$, and $\lambda_{\text{fourier}}$ on BCI, with
  $\lambda_{\text{DAB}}$ held at $0.1$. Bold marks the setting used throughout this work. These
  runs use a shortened $100$-epoch screening schedule and do not apply the additional reference
  patches used at inference, so their values are not the ones reported in
  Section~\ref{sec:experiments}. KID is reported raw here, not scaled by $10^{3}$.}
  \label{tab:weight_grid}
  \footnotesize
  \setlength{\tabcolsep}{6pt}
  \begin{tabular}{ccccccc}
    \toprule
    $\alpha$ & $\lambda_{\text{cocycle}}$ & $\lambda_{\text{fourier}}$
      & FID$\downarrow$ & KID (raw)$\downarrow$ & LPIPS$\downarrow$ & DISTS$\downarrow$ \\
    \midrule
    0.1 & 0.1 & 0.1 &  51.3954 & 0.0162 & 0.5382 & 0.2776 \\
    0.1 & 0.1 & 0.5 &  62.7132 & 0.0284 & 0.4825 & 0.2386 \\
    0.1 & 0.1 & 1.0 &  49.5539 & 0.0086 & 0.5203 & 0.2624 \\
    0.1 & 0.5 & 0.1 &  89.4415 & 0.0381 & 0.6008 & 0.3143 \\
    0.1 & 0.5 & 0.5 & 112.6470 & 0.0590 & 0.5439 & 0.2784 \\
    0.1 & 0.5 & 1.0 &  50.8993 & 0.0123 & 0.4820 & 0.2323 \\
    0.1 & 1.0 & 0.1 & 164.0038 & 0.1653 & 0.6181 & 0.3255 \\
    0.1 & 1.0 & 0.5 &  54.7145 & 0.0190 & 0.4870 & 0.2415 \\
    0.1 & 1.0 & 1.0 &  63.3100 & 0.0236 & 0.4866 & 0.2452 \\
    \midrule
    0.5 & 0.1 & 0.1 &  66.5788 & 0.0233 & 0.5143 & 0.2644 \\
    0.5 & 0.1 & 0.5 &  53.9917 & 0.0204 & 0.4948 & 0.2419 \\
    0.5 & 0.1 & 1.0 &  54.6853 & 0.0152 & 0.4856 & 0.2377 \\
    0.5 & 0.5 & 0.1 &  61.5262 & 0.0223 & 0.4860 & 0.2548 \\
    0.5 & 0.5 & 0.5 & 107.9636 & 0.0680 & 0.4920 & 0.2608 \\
    0.5 & 0.5 & 1.0 &  78.5008 & 0.0389 & 0.4912 & 0.2586 \\
    0.5 & 1.0 & 0.1 &  63.6867 & 0.0228 & 0.5540 & 0.2811 \\
    0.5 & 1.0 & 0.5 &  75.8298 & 0.0296 & 0.5361 & 0.2814 \\
    0.5 & 1.0 & 1.0 &  45.2142 & 0.0093 & 0.4772 & 0.2261 \\
    \midrule
    1.0 & 0.1 & 0.1 &  52.6085 & 0.0153 & 0.5327 & 0.2748 \\
    \textbf{1.0} & \textbf{0.1} & \textbf{0.5}
                  & \textbf{46.6457} & \textbf{0.0123} & \textbf{0.4816} & \textbf{0.2270} \\
    1.0 & 0.1 & 1.0 &  54.7589 & 0.0199 & 0.4985 & 0.2418 \\
    1.0 & 0.5 & 0.1 &  47.9143 & 0.0121 & 0.5311 & 0.2707 \\
    1.0 & 0.5 & 0.5 &  54.2129 & 0.0160 & 0.5151 & 0.2592 \\
    1.0 & 0.5 & 1.0 &  43.7119 & 0.0082 & 0.4978 & 0.2458 \\
    1.0 & 1.0 & 0.1 &  53.9382 & 0.0188 & 0.5475 & 0.2862 \\
    1.0 & 1.0 & 0.5 &  51.1449 & 0.0145 & 0.5528 & 0.2786 \\
    1.0 & 1.0 & 1.0 &  70.1446 & 0.0287 & 0.5422 & 0.2811 \\
    \bottomrule
  \end{tabular}
\end{table}

The axis with the largest effect is the pixel sheaf balance $\alpha$. Eight of the nine settings
at $\alpha = 1$ land in a narrow band of FID $43.7$ to $54.8$, and the ninth, which places
$\alpha$, $\lambda_{\text{cocycle}}$, and $\lambda_{\text{fourier}}$ all at $1.0$, reaches $70.1$.
Below $\alpha = 1$, only $7$ of $18$ settings reach that band and the worst reaches FID $164.0$.
Inside the $\alpha = 1$ region the remaining weights move the result much less. Averaged over the
grid, $\lambda_{\text{cocycle}} = 0.1$ gives the lowest mean FID of its three values, $54.8$
against $71.9$ and $71.3$.

The selected setting is second of the $27$ on mean rank over the four metrics and best on none of
them. It is first on DISTS among the $\alpha = 1$ settings, and its gap to the per-metric grid
best is $0.4\%$ on DISTS and $0.9\%$ on LPIPS, inside the range where settings are not
distinguishable. We therefore read the outcome as depending on the structure of the objective and
on $\alpha$ rather than on a precise weighting of the auxiliary terms. One setting is used
everywhere in this work, and BCI and all four MIST markers share it with no per-dataset retuning.

\paragraph{Inherent restriction map learning.}
Training under Eq.~\ref{eq:total_loss} with sheaf-consistent conditioning produces
a generator that inherently acquires the function of restriction maps. 
Gradient descent shapes $G_\theta$ into a mapping from imperfect presheaf conditioning 
to approximately sheaf-consistent output,
encoded in network parameters rather than computed by an external module.
At inference, this means independently translated adjacent patches are consistent
at their boundaries without explicit inter-patch communication,
enabling efficient parallel inference (Section~\ref{sec:pipeline}).

To test whether SheafStain learns inherent restriction-map-like behavior, 
rather than only suppressing seam artifacts at inference, 
we track the sheaf-Laplacian Dirichlet energy of the trained generator 
across its entire training trajectory.

For an input H\&E image, let $\mathcal{P}_{\text{ref}}$ denote the reference
patch grid, $G_i$ the generator output on patch $P_i$, and
$U_{ij} = P_i \cap P_j$ the overlap region of an overlapping pair. 
The image-level energy is
\begin{equation}
  E(G) \;=\; \frac{1}{|\mathcal{E}|}
  \sum_{(i,j) \in \mathcal{E}}
  \frac{1}{|U_{ij}|}\,
  \bigl\| G_i|_{U_{ij}} - G_j|_{U_{ij}} \bigr\|_2^2,
\label{eq:sheaf_dirichlet_energy}
\end{equation}
where $\mathcal{E}$ indexes adjacent-pair edges of the cover graph. 
$E(G)$ is a per-edge, per-pixel mean of the Dirichlet energy associated with
the sheaf Laplacian $\Delta_0$ of Appendix~\ref{app:sheaf}. The $1/|\mathcal{E}|$
and $1/|U_{ij}|$ factors normalize for cross-image comparison and preserve the
zero locus of the unnormalized form.
A generator that inherently respects the cocycle condition drives $E(G) \to 0$,
while a generator without spatial consistency accumulates pairwise disagreement 
on overlap regions.

We use the MIST HER2 validation set ($n = 1{,}000$ stitched H\&E images at
$1024 \times 1024$). Reference patches are sampled on a stride-$192$ grid of
$256 \times 256$ patches, producing $25$ patches and $72$ overlapping pairs
per image ($72{,}000$ pairs in total). UNSB and SheafStain on MIST HER2 are compared,
and $E(G)$ is recomputed on eight checkpoints sampled every $50$ epochs 
from epoch~$50$ to epoch~$400$.

\paragraph{Inference consistency.}
Each patch is translated independently given its cropped conditioning $M$
and per-patch neighborhood CLS token $c_n$.
Because the generator has internalized restriction maps during training, 
independently generated adjacent patches are consistent at their boundaries
even though no explicit inter-patch communication occurs at inference time,
enabling efficient parallel inference without sacrificing spatial coherence.

\section{Implementation Details}
\label{app:implementation}

\begin{figure}[t]
  \centering
  \includegraphics[width=\linewidth]{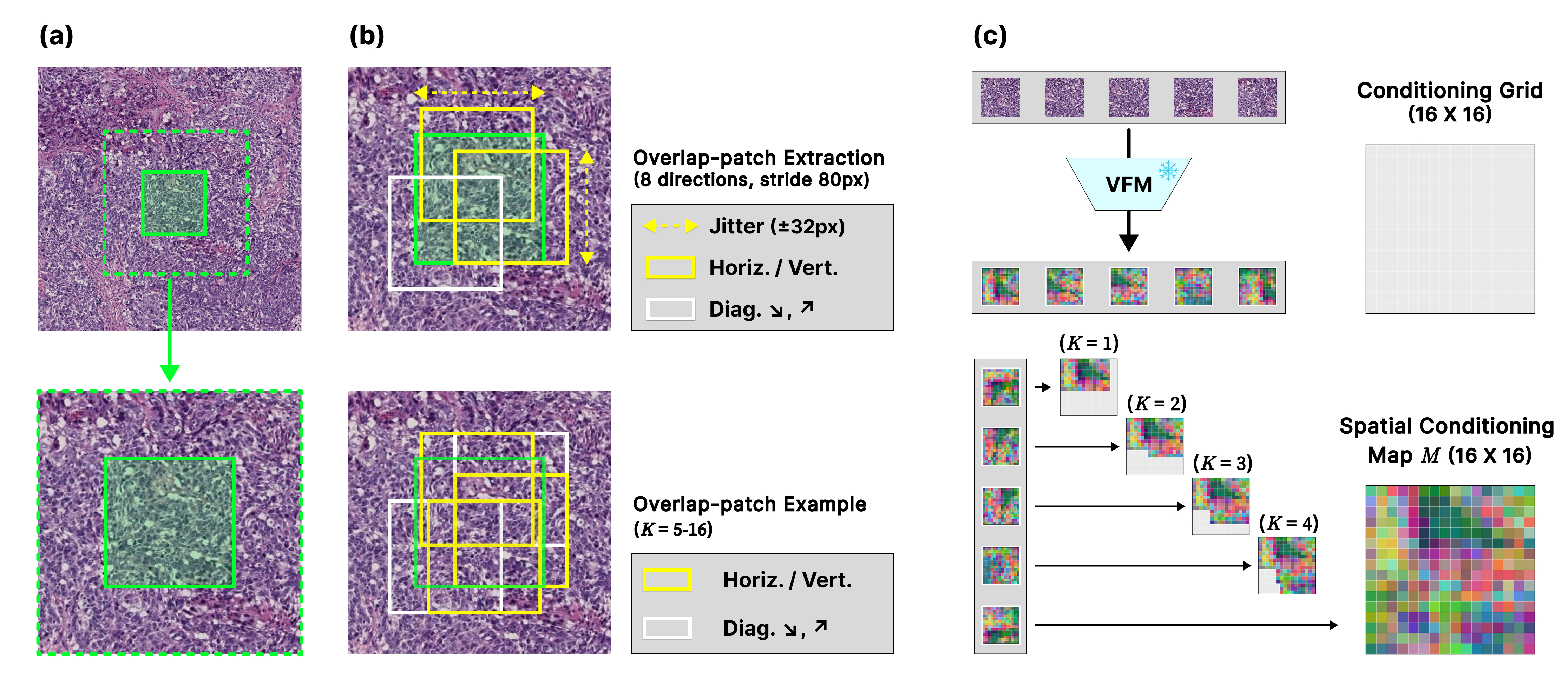}
  \caption{\textbf{Spatial map conditioning extraction pipeline in Section~\ref{sec:conditioning}}. 
  \textbf{(a)} A reference patch ($256 \times 256$) is randomly sampled from a $1024 \times 1024$ image. 
  \textbf{(b)} Surrounding overlapping patches ($224 \times 224$) are extracted in 8 directions
  at stride 80, two along each of the horizontal, vertical, and two diagonal axes
  (lower-left to upper-right, upper-left to lower-right). 
  Horizontal and vertical directions receive lateral jitter ($\pm 32$\,px) 
  and contribute one center and two jittered patches each, while diagonal directions contribute 
  one patch each, yielding 5--16 patches per reference. 
  \textbf{(c)} Each surrounding patch's $14{\times}14$ patch tokens 
  are mapped onto the reference's $16{\times}16$ conditioning grid by floor-rounded offset. 
  Tokens contributing to the same cell are averaged, producing the spatial conditioning map $M$.}
  \label{fig:pipeline}
\end{figure}

\paragraph{Neighborhood construction.}
For a reference patch $P_{\text{ref}}$ ($256 \times 256$) to be translated, 
we extract $K$ surrounding overlapping patches ($224 \times 224$) in 8 directions
at stride 80 (Figure~\ref{fig:pipeline}). Two directions lie along each of the horizontal axis, the vertical axis,
and the two diagonal axes (lower-left to upper-right, upper-left to lower-right). 
The horizontal and vertical directions additionally receive lateral jitter ($\pm 32$\,px) 
to maximize boundary coverage and contribute one center and two jittered patches each. 
The diagonal directions contribute one patch each, yielding $K \in [5, 16]$ per reference 
depending on tissue coverage.

\paragraph{Token-to-grid alignment.}
The reference patch's $16 \times 16$ conditioning grid corresponds
to pixel coordinates $(16y + 8,\, 16x + 8)$ for $(y, x) \in [0, 16)^2$. 
Each surrounding $224 \times 224$ patch produces $14 \times 14$ tokens at analogous positions 
within its own coordinate system. 
A patch offset by $(d_x, d_y)$ pixels relative to the reference patch maps
its token at patch-local position $(y', x')$ to reference-local grid cell 
$(y' + \lfloor d_y / 16 \rfloor,\, x' + \lfloor d_x / 16 \rfloor)$, 
retained when the mapped cell lies within $[0, 16)^2$. 
When jitter makes $(d_x, d_y)$ a non-multiple of $16$, the token contributes 
to its nearest-integer grid cell. Multiple tokens contributing to the same cell are averaged. 
Cells with no contributing token (rare under 8-direction coverage with jitter) default 
to zero. Figure~\ref{fig:pipeline}(c) illustrates the mapping.

\FloatBarrier
\section{Inference-Time Sheafification}
\label{app:inference}

This section details the inference-time construction summarized in
Section~\ref{sec:pipeline} and used to produce all reported SheafStain results.
Throughout we work on a $1024 \times 1024$ image $X$.

\paragraph{Open cover.}
The cover $\mathcal{U} = \{U_i\}_{i=0}^{N-1}$ of $256 \times 256$ patches is
assembled in two stages. The first stage places reference patches on a uniform
stride-$192$ grid with the trailing row and column snapped to the image boundary,
yielding $25$ patches whose union covers $X$ with $64$-pixel overlap between
stride-adjacent neighbors. The second stage adds $9$ additional reference patches by
computing a mid-band FFT energy map ($f \in [1/16, 1/4]$ cycles/pixel) over the
$25$-patch lattice and greedily placing patches at energy extremes
(priority $2|\mathrm{rank}(\text{energy}) - 1/2|$) under a $112$\,px
minimum-distance constraint. From the sheaf-theoretic perspective, a finer cover tightens
the \v{C}ech approximation, and the bimodal heuristic densifies the cover where
sheafification error concentrates. In total, the cover contains $N = 34$ patches.

\begin{figure}[t]
  \centering
  \includegraphics[width=\linewidth]{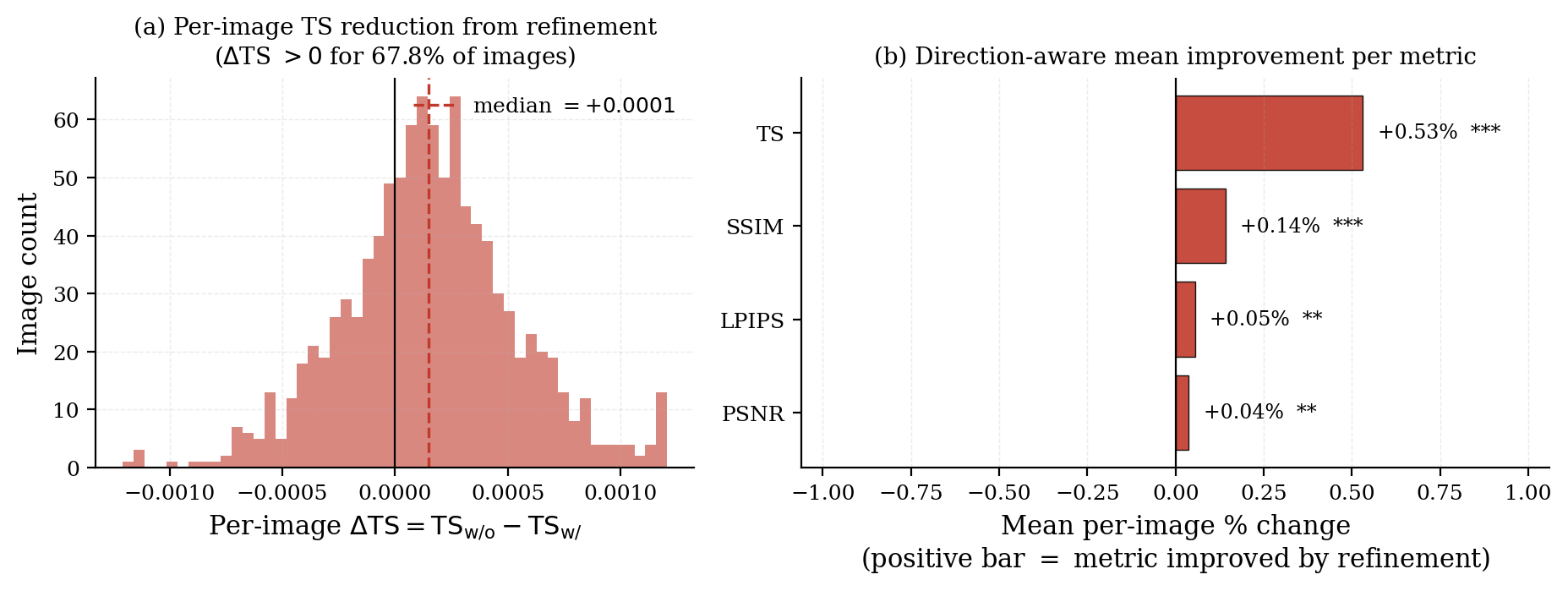}
  \caption{\textbf{Effect of the additional reference patches on MIST HER2
  validation ($n = 1{,}000$, stride $192$, $256 \times 256$ reference
  patches).} \textbf{(a)} Histogram of per-image
  $\Delta\mathrm{TS} = \mathrm{TS}_{\text{w/o}} - \mathrm{TS}_{\text{w/}}$.
  Bars to the right of zero are images for which adding the patches
  reduced TS. \textbf{(b)} Mean per-image \% change for four paired
  metrics, with the sign flipped on higher-is-better metrics so that a
  positive bar always denotes improvement. Significance from paired
  Wilcoxon tests:
  $^{\ast}\,p<5{\times}10^{-2}$,
  $^{\ast\ast}\,p<10^{-3}$,
  $^{\ast\ast\ast}\,p<10^{-10}$.}
  \label{fig:refinement_ablation_mist}
\end{figure}

\paragraph{Per-patch conditioning.}
For each $U_i$, the spatial map $M_i$ and neighborhood CLS token $c_i$ are
constructed by the same VFM scan over overlapping patches and token-to-grid scatter
described in Appendix~\ref{app:implementation}. At inference, the number of
overlapping patches $K_i$ is coupled to local complexity. Reference patches
use $K_i \in [5, 16]$ via linear interpolation between image-wide $25$th
and $75$th percentiles of FFT energy. This mirrors training-time per-reference variability of $K$
and preserves training-inference symmetry.

\paragraph{Stitching as partition-of-unity gluing.}
With per-patch outputs $\{s_i\}_{i=0}^{N-1}$ from independent generator
forwards, the final image is the normalized weighted sum
\begin{equation}
  \hat{Y}(x, y) = \frac{\sum_{i \,:\, (x,y) \in U_i} w_i(x, y)\, s_i(x, y)}
                       {\sum_{i \,:\, (x,y) \in U_i} w_i(x, y)},
  \label{eq:pou_stitch}
\end{equation}
with two compactly-supported kernels:
\begin{itemize}
  \item \emph{Reference patch} ($i < 25$): $w_i$ is a separable linear ramp
    rising from $0$ to $1$ over the $32$-pixel border and equal to $1$
    on the central $192 \times 192$ plateau.
  \item \emph{Additional reference patch} ($i \geq 25$): $w_i$ is a separable raised-cosine
    taper, peaking at the patch center and smoothly decaying to $0$ at
    the boundary (no central plateau).
\end{itemize}
Normalization converts $\{w_i\}$ into a subordinate partition of unity
$\phi_i := w_i / \sum_j w_j$ satisfying $\sum_i \phi_i \equiv 1$, recovering
the standard sheaf gluing form $\hat{Y} = \sum_i \phi_i s_i$. Reference-patch
coverage ensures the denominator is strictly positive everywhere. The
hybrid choice is deliberate. A uniform additional-patch kernel would introduce
sharp seams at additional-patch boundaries, whereas the cosine taper vanishes there
and joins the reference-patch weighting smoothly.

By construction, training and inference share the same \v{C}ech data.
Training drives $\|\delta s\|^2 \to 0$ on overlaps via the pixel sheaf
and cocycle losses (Section~\ref{sec:sheaf_loss}), and inference assembles
generator outputs through a smooth subordinate partition of unity over
the same cover. Any residual overlap disagreement is therefore absorbed by the
convex average rather than left as a visible seam.

\begin{figure}[t]
  \centering
  \includegraphics[width=0.8\linewidth]{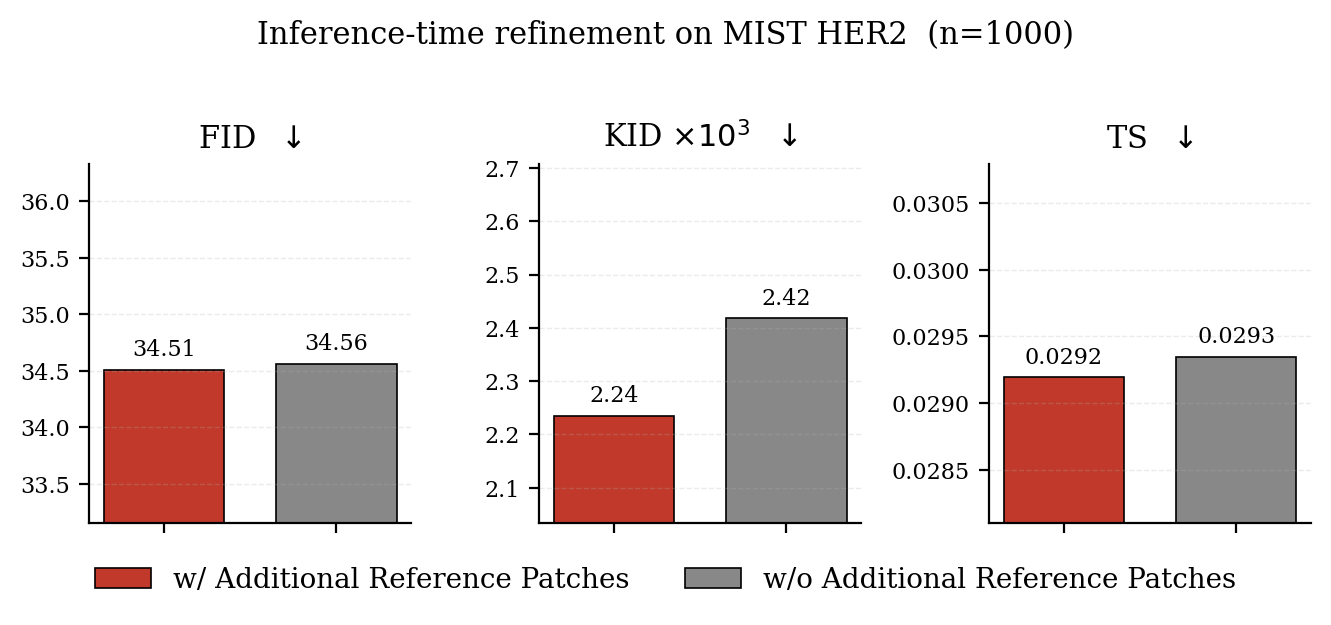}
  \caption{\textbf{Dataset-level metric comparison on MIST HER2 validation
  ($n = 1{,}000$, stride $192$)}. Each panel uses an independent y-axis
  to make the with/without contrast visible at the metric's natural scale.
  KID is reported as $\mathrm{KID} \times 10^{3}$ for legibility.
  Adding the patches lowers FID, KID, and TS.}
  \label{fig:refinement_metrics_mist}
\end{figure}

\paragraph{Effect of the additional reference patches.}
On the MIST HER2 validation split ($n = 1{,}000$, stride $192$),
the additional reference patches improve every paired
metric we report
(Figure~\ref{fig:refinement_ablation_mist}). Panel~(a) plots the
per-image shift
$\Delta\mathrm{TS} = \mathrm{TS}_{\text{w/o}} - \mathrm{TS}_{\text{w/}}$.
The additional reference patches reduce TS on $678$ of $1{,}000$ images
($67.8\%$, paired Wilcoxon $p < 10^{-33}$). Panel~(b) reports the mean
per-image \% change for four paired metrics with the sign flipped on
higher-is-better metrics, so that a positive bar always denotes
improvement. TS, SSIM, LPIPS, and PSNR are all improved at
$p < 0.05$.
Figure~\ref{fig:refinement_metrics_mist} adds the dataset-level view.
The additional reference patches lower FID, KID, and TS, so the refinement
improves perceptual and seam metrics at the dataset level as well.

\FloatBarrier
\section{Dataset Statistics}
\label{app:datasets}

Table~\ref{tab:dataset_app} reports paired image counts per dataset and IHC marker, 
supporting the split description in Section~\ref{sec:setup}. 
BCI provides a single HER2 partition with official train/test splits,
while MIST provides train/validation splits for each of the four biomarkers.

\begin{table}[h]
  \caption{Paired image counts per dataset and IHC marker.}
  \label{tab:dataset_app}
  \centering
  \setlength{\tabcolsep}{10pt}
  \begin{tabular}{llrr}
    \toprule
    Dataset & Stain & Train & Valid/Test \\
    \midrule
    BCI~\citep{liu2022bci}  & HER2  & 3,896 & 977 \\
    \midrule
    \multirow{4}{*}{MIST~\citep{mist}}
                            & HER2  & 4,642 & 1,000 \\
                            & Ki-67 & 4,361 & 1,000 \\
                            & ER    & 4,153 & 1,000 \\
                            & PR    & 4,139 & 1,000 \\
    \bottomrule
  \end{tabular}
\end{table}

\paragraph{Dataset usage and evaluation splits.} 
We utilize the official training partitions of both BCI and MIST datasets
for training SheafStain and all comparison methods. 
Regarding performance evaluation, we employ the official BCI test set 
as our held-out benchmark. 
Since the MIST dataset lacks an official test partition, 
we repurpose its validation sets to evaluate and report performance 
across the four IHC biomarkers (HER2, Ki-67, ER, and PR).

\section{Empirical Verification of Sheaf \texorpdfstring{Axioms}{Axioms}}
\label{app:gluing}

This section expands the compressed verification of
Section~\ref{sec:sheaf}, providing the formal mechanism behind context
contamination, its scope across encoder families, and full empirical
evidence on VFM embeddings.

\paragraph{Setup.}
Let $X$ denote a histopathology image and $\mathcal{U} = \{U_i\}$ an
open cover of $X$ by overlapping patches. The frozen VFM $\Phi$ maps
each patch $U_i$ to a collection of token embeddings
$\Phi(U_i) = \{\Phi(U_i)_t\}_{t=1}^{T} \in \mathbb{R}^d$
($T = 196$, $d = 1536$), spatially arranged and
consistently indexed across overlapping patches. Let
$s_i := \Phi(U_i) \in \sheafF(U_i)$ denote the section produced by
the VFM on $U_i$. The restriction map
$\rho_{U_i \to U_{ij}} : \sheafF(U_i) \to \sheafF(U_{ij})$ is defined
by spatial indexing, taking $s_i$ to the sub-collection of tokens
whose positions lie in the overlap $U_{ij}$. This assignment defines
a presheaf $\sheafF$ on $X$.

\paragraph{Locality (verified).}
Token extraction is purely positional. Restricting $s_i$ directly
to a sub-region yields identical tokens to first restricting to an
intermediate region and then restricting further (max difference
$= 0$), confirming that the restriction maps compose as a presheaf
requires. Locality itself is immediate, since a section is determined by its
token positions.

\paragraph{Gluing (compatibility fails).}
For adjacent patches $U_i, U_j$ sharing an overlap region $U_{ij}$,
the compatibility that gluing requires (Eq.~\ref{eq:compatibility}) is
$\rho_{U_i \to U_{ij}}(s_i) = \rho_{U_j \to U_{ij}}(s_j)$. Measuring
cosine similarity between corresponding overlap tokens reveals a
systematic gap. On H\&E, the mean cardinal cosine ranges from $0.63$ (stride 192,
$14\%$ overlap) to $0.92$ (stride 32, $86\%$ overlap), and cosine stays
below the ideal $1.0$ across all strides, directions, and stain types
(Table~\ref{tab:gluing}). These sections thus fail to satisfy the
compatibility that gluing requires, so no single global section restricts to them. The Dirichlet energy
$E(x) = \langle \Delta_0 x, x \rangle > 0$ is strictly positive, 
quantifying the global inconsistency that manifests as stitching artifacts when
translated patches are reassembled.

\paragraph{Empirical evidence of context contamination.}
As formulated in Eq.~\ref{eq:context_full}, the same token receives
different non-overlapping context depending on its parent patch. 
This failure of the overlap compatibility that gluing requires is not a dataset-specific artifact
but an inherent byproduct of the attention mechanism.
This is empirically supported by the near-identical magnitude of the gap 
between H\&E and IHC ($\Delta < 0.02$ across all strides), 
which confirms the stain-independence of this phenomenon 
and justifies the role allocation of the VFM in SheafStain.

\paragraph{Scope of context contamination.}
The mechanism analyzed in Eq.~\ref{eq:context_full} is specific to
\emph{global} self-attention. Encoders with strictly local receptive
fields (e.g., pure CNNs) would produce approximately compatible sections
by construction, and encoders with restricted attention windows
(e.g., windowed Swin) would exhibit reduced contamination. In
practice, however, state-of-the-art pathology VFMs are predominantly
ViT-based with global attention (Prov-GigaPath~\citep{gigapath}, UNI~\citep{uni}, Virchow2~\citep{virchow}),
making context contamination a structural concern for the current
pathology VFM ecosystem. The sheaf losses proposed in
Section~\ref{sec:sheaf_loss} are encoder-agnostic and automatically
scale with the residual compatibility gap of whichever encoder is chosen.

\paragraph{Implications for patch-wise virtual staining.}
In patch-wise translation with the baseline, each patch is translated
independently. The generator $G$ receives a noisy input $x_t$ and
produces $G(x_t, t, z)$ without knowledge of adjacent patches. The
resulting translated patches inherit the presheaf structure,
being locally valid but globally incoherent. Rather than correcting VFM
embeddings via an external module, SheafStain injects spatial context from
surrounding patches as conditioning and enforces the compatibility that gluing requires
through explicit losses, allowing the generator to
\emph{inherently learn restriction maps} that produce globally
consistent outputs (Appendix~\ref{app:training}).

\paragraph{Stride selection rationale.}
We measure overlap-token consistency on BCI at eight strides
$\{32, 64, 80, 96, 112, 128, 160, 192\}$. Three of these
$(32, 80, 160)$ satisfy $(1024 - 224)/s \in \mathbb{Z}$ and
therefore cover the $1024 \times 1024$ image without boundary
crop, yielding integer-aligned grids of $26^2$, $11^2$, and
$6^2$ patches respectively. The remaining five strides
$(64, 96, 112, 128, 192)$ densely sample the overlap-ratio
axis between $14\%$ and $71\%$, characterizing the continuous
dependence of cosine similarity on overlap.

\paragraph{Empirical magnitude.}
Table~\ref{tab:gluing} reports mean cosine similarity across
all eight strides and four directions on both H\&E and IHC.
The signal is monotone. At $86\%$ overlap (stride 32), H\&E
cardinal cosine reaches $0.918$--$0.920$, declining smoothly
through intermediate strides to $0.626$--$0.628$ at $14\%$
overlap (stride 192). Diagonals follow the same pattern at
consistently lower magnitude (corner overlaps contain fewer
non-boundary tokens than rectangular ones). Mean L2 distances
follow the inverse pattern, ranging from $\approx 11.7$ at
stride 32 to $\approx 26.0$ at stride 192 for cardinal
directions, and from $\approx 16.0$ to $\approx 29.8$ for
diagonals.

\begin{table}[h]
  \caption{Mean cosine similarity of overlap tokens across all eight strides 
           and four directions (4{,}873 BCI images, $1024 \times 1024$). 
           Underlined strides evenly cover $1024 \times 1024$ without boundary crop.
           Ideal sheaf consistency is $1.0$. 
           Cosine remains below the ideal across all strides, directions, and stains.}
  \label{tab:gluing}
  \centering
  \small
  \setlength{\tabcolsep}{4pt}
  \begin{tabular}{rrcccc|cccc}
    \toprule
    \multirow{2}{*}{Stride} & \multirow{2}{*}{Overlap}
    & \multicolumn{4}{c}{H\&E} & \multicolumn{4}{c}{IHC} \\
    \cmidrule(lr){3-6} \cmidrule(lr){7-10}
    & & Horizontal & Vertical & Diag. $\searrow$ & Diag. $\nearrow$ & Horizontal & Vertical & Diag. $\searrow$ & Diag. $\nearrow$ \\
    \midrule
    \underline{32}  & 85.7\% & 0.918 & 0.920 & 0.856 & 0.856 & 0.910 & 0.913 & 0.846 & 0.846 \\
    64              & 71.4\% & 0.849 & 0.853 & 0.748 & 0.748 & 0.837 & 0.844 & 0.736 & 0.735 \\
    \underline{80}  & 64.3\% & 0.824 & 0.829 & 0.716 & 0.717 & 0.810 & 0.818 & 0.703 & 0.702 \\
    96              & 57.1\% & 0.802 & 0.806 & 0.690 & 0.691 & 0.787 & 0.794 & 0.676 & 0.675 \\
    112             & 50.0\% & 0.781 & 0.786 & 0.668 & 0.669 & 0.765 & 0.772 & 0.653 & 0.653 \\
    128             & 42.9\% & 0.753 & 0.757 & 0.635 & 0.636 & 0.738 & 0.745 & 0.623 & 0.622 \\
    \underline{160} & 28.6\% & 0.694 & 0.698 & 0.569 & 0.571 & 0.680 & 0.687 & 0.562 & 0.562 \\
    192             & 14.3\% & 0.626 & 0.628 & 0.500 & 0.504 & 0.613 & 0.618 & 0.494 & 0.494 \\
    \bottomrule
  \end{tabular}\\[3pt]
  {\footnotesize Diag. $\searrow$: upper-left to lower-right diagonal axis. 
                 Diag. $\nearrow$: lower-left to upper-right diagonal axis.}
\end{table}

\paragraph{Cardinal vs.\ diagonal geometry.}
At every stride the horizontal and vertical directions are
nearly indistinguishable. Their cosine values differ by at
most $0.008$ on both H\&E and IHC across all eight strides,
consistent with the four-fold rotational symmetry of ViT
self-attention applied to square inputs. The two diagonal
axes (upper-left to lower-right, and lower-left to
upper-right) are likewise indistinguishable from each other
($\Delta \le 0.004$) but lag the cardinal directions by
$0.06$--$0.13$ in cosine, reflecting the smaller
non-boundary-token mass in corner-shaped overlaps.

\begin{figure}[!htbp]
  \centering
  \includegraphics[width=\linewidth]{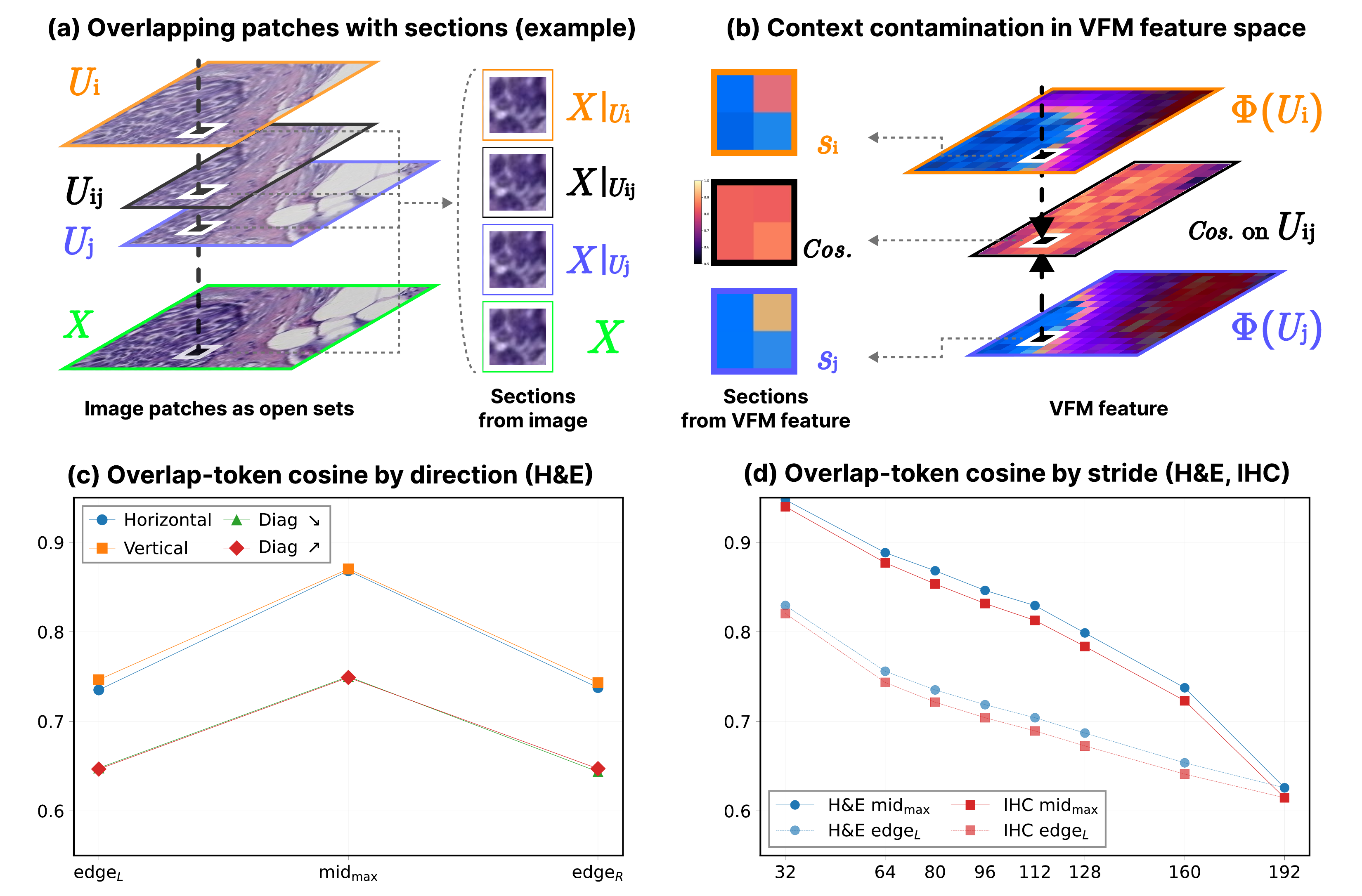}
  \caption{\textbf{Per-position analysis of Prov-GigaPath overlap-token consistency on 4{,}873 BCI images.} 
           \textbf{(a)} Two adjacent $224 \times 224$ patches $U_i, U_j$ of an example BCI image $X$, cut at
           stride 80, are open sets that share an overlap $U_{ij}$. The image itself
           gives a section over each of them, its restriction $X|_{U_i}$, $X|_{U_{ij}}$, or $X|_{U_j}$, and
           these agree at every position of the overlap, so they glue into $X$.
           \textbf{(b)} Context contamination in the frozen VFM. The sections $s_i = \Phi(U_i)$ and
           $s_j = \Phi(U_j)$ from Prov-GigaPath are colored by a PCA fitted on the tokens of both patches,
           so matching colors mean similar embeddings. At the same position in $U_{ij}$ the two sections
           disagree, because each patch adds its own non-overlapping context. The cosine similarity map on $U_{ij}$
           compares them position by position. Its mean is $0.80$, and it is lower at the strip boundary ($0.73$)
           than in its interior (up to $0.86$). The dashed line marks one position $t$ in $U_{ij}$, and the insets
           in (a) and (b) show each section around $t$.
           \textbf{(c)} At stride 80, mean cosine similarity at the boundary positions 
           of the overlap strip ($\mathrm{edge}_L, \mathrm{edge}_R$) is consistently lower than 
           the maximum cosine across middle positions ($\mathrm{mid}_{\max}$), forming the inverted-U pattern.
           The gap is $\approx 0.13$ for cardinal directions and $\approx 0.10$ for diagonals. 
           \textbf{(d)} Boundary cosine ($\mathrm{edge}_L$, dashed) and middle cosine 
           ($\mathrm{mid}_{\max}$, solid) for the horizontal direction across all eight strides, 
           with H\&E (blue) and IHC (red) overlaid. Both metrics decline with stride, 
           the inverted-U gap persists through stride 160, and the H\&E/IHC pair tracks within $\Delta < 0.02$ 
           at every stride.}
  \label{fig:gluing_perposition}
\end{figure}

\paragraph{Per-position analysis.}
Within an overlap strip, tokens at the strip boundary inherit
more non-shared context than tokens at the strip interior.
Their attention windows extend further into the
non-overlapping region of one of the two patches. This
predicts a systematic boundary-vs-interior gap.
Table~\ref{tab:per_position} reports cosine at the two
boundary positions of the overlap strip
($\mathrm{edge}_L, \mathrm{edge}_R$) and the maximum cosine
across middle positions ($\mathrm{mid}_{\max}$) for strides
80 and 160. The
$\mathrm{mid}_{\max} \!>\! \mathrm{edge}$ gap is consistent,
at $\approx 0.13$ for stride 80 and $\approx 0.08$ for
stride 160 (cardinal).
Figure~\ref{fig:gluing_perposition}(a, b) shows the
mechanism on an example BCI image. Two adjacent patches share
an overlap region (a), and after the frozen VFM the tokens at
the same position in that region differ between $U_i$ and
$U_j$, because each is attended over a different
non-overlapping context (b, context contamination). The
example already shows the boundary-vs-interior gap, with a
cosine of $0.73$ at the strip boundary against up to $0.86$
inside. Figure~\ref{fig:gluing_perposition}(c)
visualizes the resulting inverted-U pattern at stride 80
across all four directions. Figure~\ref{fig:gluing_perposition}(d)
overlays H\&E and IHC and shows that both
$\mathrm{edge}_L$ and $\mathrm{mid}_{\max}$ decline with
stride while the gap between them persists through stride 160, indicating that the
contamination ratio scales with overlap independently of
position bias. The same edge position $k{=}0$ at
different strides yields different cosine
($0.735$ at stride 80 vs.\ $0.654$ at stride 160 on H\&E
horizontal, $\Delta = 0.081$), isolating contamination ratio
from position bias.

\begin{table}[h]
  \caption{Per-position cosine similarity in the overlap strip 
           at strides 80 and 160 (4{,}873 BCI images). 
           The systematic $\mathrm{mid}_{\max} \!>\! \mathrm{edge}$ gap 
           (inverted-U) confirms that boundary tokens agree less than 
           central overlap tokens, as predicted by 
           the context-contamination mechanism (Eq.~\ref{eq:context_full}).}
  \label{tab:per_position}
  \centering
  \small
  \setlength{\tabcolsep}{4pt}
  \begin{tabular}{rlrccc|ccc}
    \toprule
    \multirow{2}{*}{Stride} & \multirow{2}{*}{Direction} & \multirow{2}{*}{Ovl. tokens}
    & \multicolumn{3}{c|}{H\&E} & \multicolumn{3}{c}{IHC} \\
    \cmidrule(lr){4-6} \cmidrule(lr){7-9}
    & & & $\mathrm{edge}_L$ & $\mathrm{mid}_{\max}$ & $\mathrm{edge}_R$ & $\mathrm{edge}_L$ & $\mathrm{mid}_{\max}$ & $\mathrm{edge}_R$ \\
    \midrule
    \multirow{4}{*}{80}
      & Horizontal        & 9 & 0.735 & 0.868 & 0.738 & 0.721 & 0.854 & 0.724 \\
      & Vertical          & 9 & 0.747 & 0.870 & 0.743 & 0.739 & 0.856 & 0.737 \\
      & Diag. $\searrow$  & 9 & 0.648 & 0.750 & 0.644 & 0.638 & 0.734 & 0.636 \\
      & Diag. $\nearrow$  & 9 & 0.647 & 0.749 & 0.647 & 0.635 & 0.731 & 0.637 \\
    \midrule
    \multirow{4}{*}{160}
      & Horizontal        & 4 & 0.654 & 0.738 & 0.651 & 0.641 & 0.723 & 0.638 \\
      & Vertical          & 4 & 0.660 & 0.739 & 0.655 & 0.651 & 0.726 & 0.647 \\
      & Diag. $\searrow$  & 4 & 0.539 & 0.602 & 0.534 & 0.532 & 0.594 & 0.529 \\
      & Diag. $\nearrow$  & 4 & 0.537 & 0.605 & 0.537 & 0.529 & 0.595 & 0.528 \\
    \bottomrule
  \end{tabular}\\[3pt]
  {\footnotesize Ovl. tokens: number of token positions across the overlap strip. 
                  $\mathrm{edge}_L, \mathrm{edge}_R$: leftmost/rightmost overlap positions. 
                  $\mathrm{mid}_{\max}$: maximum cosine across middle positions. 
                  Direction labels as in Table~\ref{tab:gluing}.}
\end{table}

\paragraph{Stability across H\&E and IHC.}
The maximum H\&E/IHC discrepancy across all $32$ stride $\times$ direction combinations
is $|\Delta| = 0.017$ (stride 112, lower-left to upper-right diagonal),
the mean discrepancy is $0.011$, and all combinations satisfy $|\Delta| < 0.02$.
This near-equality shows that the encoder's overlap behavior does not depend on the stain. 
The same encoder supplies morphological conditioning ($M$, $c_{n}$ from H\&E patches)
and VFM Alignment ($\mathcal{L}_{\text{stain-align}}$) on DAB-isolated renderings,
which inherits the VFM's IHC-side coverage of chromogen patterns.

\begin{figure}[!htbp]
  \centering
  \begin{subfigure}[t]{0.48\linewidth}
    \centering
    \includegraphics[width=\linewidth]{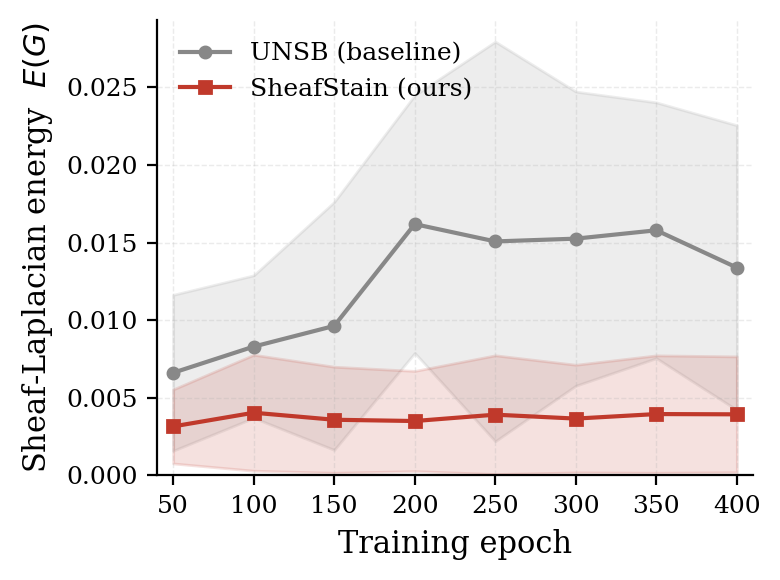}
    \caption{\textbf{Sheaf-Laplacian Dirichlet energy across training.} Shaded bands: $\pm 1$ per-image std.}
    \label{fig:sheaf_energy_traj}
  \end{subfigure}
  \hfill
  \begin{subfigure}[t]{0.48\linewidth}
    \centering
    \includegraphics[width=\linewidth]{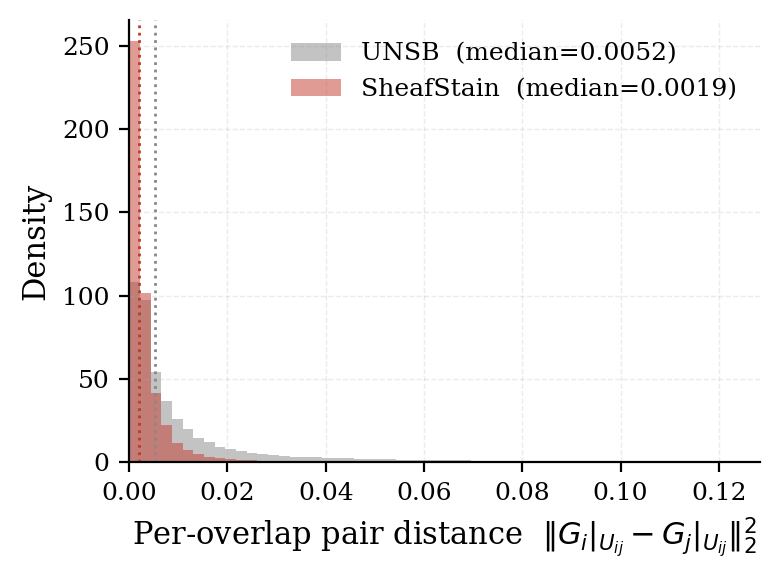}
    \caption{\textbf{Per-pair distance distribution.} Squared distances at epoch~$400$
    for SheafStain (red) and UNSB (gray). Dotted lines mark medians.}
    \label{fig:sheaf_pair_hist}
  \end{subfigure}
  \caption{Empirical verification of inherent restriction map learning on MIST HER2
  validation set. 
  \textbf{(\subref{fig:sheaf_energy_traj}) Sheaf-Laplacian Dirichlet energy
  trajectory across training.} SheafStain remains within $[0.003,\, 0.004]$ 
  from epoch~$50$ onwards. UNSB diverges with training.
   \textbf{(\subref{fig:sheaf_pair_hist}) Per-pair distance distribution.}
  Squared distances at the final epoch, pooled across $72{,}000$ overlapping pairs. 
  SheafStain concentrates near zero. UNSB has a heavier right tail.
  \label{fig:sheaf_energy_combined}}
\end{figure}

\paragraph{Energy trajectory and per-pair distribution at the final checkpoint.}
Figure~\ref{fig:sheaf_energy_traj} reports $E(G)$ as a function of training
epoch. SheafStain reaches $E \approx 0.0032$ by epoch~$50$ and remains within
the narrow band $[0.0032,\, 0.0040]$ across all subsequent checkpoints. The
per-image standard deviation also stays low ($0.0024$--$0.0038$). UNSB
starts at $E = 0.0066$ at epoch~$50$ and drifts \emph{upwards} as training
progresses, reaching a plateau of $E \approx 0.013$--$0.016$ from
epoch~$200$ onwards. The ratio between UNSB and SheafStain energies grows
from $2.1\times$ at epoch~$50$ to $4.6\times$ at epoch~$200$, and remains
$\geq 3.4\times$ for every later checkpoint. This trajectory shows that the
sheaf conditioning and consistency losses establish restriction-map-like
behavior early in training and preserve it across later epochs, while a
purely generative objective (UNSB) yields the opposite trend, in which the overlap disagreement
grows as the generator becomes more expressive.

Figure~\ref{fig:sheaf_pair_hist} reports the distribution of per-pair
squared distances at epoch~$400$, pooled across all $72{,}000$ pairs.
SheafStain concentrates sharply near zero (median $0.0019$, $95$th
percentile $0.0126$), whereas UNSB has a broader distribution with a
heavier right tail (median $0.0052$, $95$th percentile $0.0584$). The
SheafStain median is $36\%$ of the UNSB median and the $95$th percentile is
$22\%$ of UNSB, indicating that the improvement is not driven by a few
outlier pairs but is consistent across the overlap population. Together
with the trajectory above, this provides direct evidence (measured
internally to the generator, prior to any stitching post-process) that
SheafStain approximates a section of the sheaf $\mathcal{F}$, supporting the
restriction-map interpretation that motivates the framework.

\FloatBarrier
\section{Quantitative Performance Comparison}
\label{app:full_comparison}

This section expands Section~\ref{sec:baselines} and
Table~\ref{tab:comparison} with the complete 12-metric set across all
seven methods and five (dataset, stain) blocks, split into stitched image
quality (Table~\ref{tab:full_metrics_1024_a}) and biomarker quantification (Table~\ref{tab:full_metrics_1024_b}).
The $1024 \times 1024$ tables follow the protocol of
Section~\ref{sec:setup}. Pix2pix, CycleGAN, PSPStain, UNIStainNet, and UNSB use simple $4 \times 4$
row-major aggregation, whereas D-VST and SheafStain use their native inference.

\begin{table}[h]
  \caption{Full image-quality and stitching metrics at $1024 \times 1024$ across all 7 methods and 5 (dataset, stain) blocks.
           Five prior methods use simple $4 \times 4$ row-major aggregation.
           D-VST and SheafStain each follow their native inference.
           Values are means (FID and KID) or mean$\pm$std, reported to 4 decimal places.
           Best metrics in bold, second-best metrics underlined.}
  \label{tab:full_metrics_1024_a}
  \centering
  \scriptsize
  \setlength{\tabcolsep}{1.5pt}
  \resizebox{\textwidth}{!}{%
  \begin{tabular}{lllccccccc}
    \toprule
    Dataset & Stain & Method & FID$\downarrow$ & KID$\times 10^{3}\downarrow$ & LPIPS$\downarrow$ & DISTS$\downarrow$ & PSNR$\uparrow$ & SSIM$\uparrow$ & TS$\downarrow$ \\
    \midrule
    \multirow{7}{*}{BCI} & \multirow{7}{*}{HER2} & pix2pix & 172.7932 & 138.9050 & 0.4800{\scriptsize$\pm$0.0722} & 0.2753{\scriptsize$\pm$0.0698} & \textbf{21.2036{\scriptsize$\pm$4.3555}} & 0.5290{\scriptsize$\pm$0.1586} & 0.0533{\scriptsize$\pm$0.0214} \\
    & & CycleGAN & 96.3496 & 50.8020 & 0.5448{\scriptsize$\pm$0.0659} & 0.2920{\scriptsize$\pm$0.0599} & 16.4494{\scriptsize$\pm$5.0680} & \textbf{0.5433{\scriptsize$\pm$0.1571}} & 0.0475{\scriptsize$\pm$0.0262} \\
    & & PSPStain & 185.6991 & 173.7360 & 0.5695{\scriptsize$\pm$0.0640} & 0.2924{\scriptsize$\pm$0.0494} & 16.2668{\scriptsize$\pm$4.0559} & 0.5188{\scriptsize$\pm$0.1411} & 0.0885{\scriptsize$\pm$0.0354} \\
    & & D-VST & 87.3100 & 62.7670 & 0.5107{\scriptsize$\pm$0.0707} & 0.2570{\scriptsize$\pm$0.0596} & 18.4934{\scriptsize$\pm$4.4975} & 0.4855{\scriptsize$\pm$0.1864} & \underline{0.0164{\scriptsize$\pm$0.0070}} \\
    & & UNIStainNet & \underline{67.6322} & \underline{29.3000} & \textbf{0.4577{\scriptsize$\pm$0.0711}} & \underline{0.2279{\scriptsize$\pm$0.0531}} & \underline{19.9257{\scriptsize$\pm$5.2330}} & \underline{0.5379{\scriptsize$\pm$0.1775}} & 0.0448{\scriptsize$\pm$0.0320} \\
    & & UNSB & 227.7889 & 236.2310 & 0.6635{\scriptsize$\pm$0.0677} & 0.3347{\scriptsize$\pm$0.0452} & 13.3611{\scriptsize$\pm$1.8560} & 0.3850{\scriptsize$\pm$0.1370} & 0.1464{\scriptsize$\pm$0.0161} \\
    & & \textbf{SheafStain (ours)} & \textbf{36.3626} & \textbf{4.2070} & \underline{0.4689{\scriptsize$\pm$0.0584}} & \textbf{0.2132{\scriptsize$\pm$0.0479}} & 19.7809{\scriptsize$\pm$4.7203} & 0.5176{\scriptsize$\pm$0.1749} & \textbf{0.0146{\scriptsize$\pm$0.0072}} \\
    \midrule
    \multirow{28}{*}{MIST} & \multirow{7}{*}{HER2} & pix2pix & 186.3988 & 186.3800 & 0.5631{\scriptsize$\pm$0.0619} & 0.3624{\scriptsize$\pm$0.0685} & \textbf{15.0192{\scriptsize$\pm$2.7204}} & \textbf{0.2519{\scriptsize$\pm$0.0769}} & 0.0851{\scriptsize$\pm$0.0196} \\
    & & CycleGAN & 95.9094 & 54.0670 & 0.5766{\scriptsize$\pm$0.0548} & 0.2791{\scriptsize$\pm$0.0454} & 12.9409{\scriptsize$\pm$2.2645} & 0.2228{\scriptsize$\pm$0.0916} & 0.0698{\scriptsize$\pm$0.0110} \\
    & & PSPStain & \underline{50.1078} & \underline{10.6750} & \underline{0.5371{\scriptsize$\pm$0.0594}} & \underline{0.2466{\scriptsize$\pm$0.0388}} & 13.9313{\scriptsize$\pm$2.3381} & 0.2256{\scriptsize$\pm$0.0744} & 0.0763{\scriptsize$\pm$0.0210} \\
    & & D-VST & 94.0706 & 65.4310 & 0.5578{\scriptsize$\pm$0.0525} & 0.2613{\scriptsize$\pm$0.0462} & 13.3108{\scriptsize$\pm$2.2011} & 0.2049{\scriptsize$\pm$0.0612} & \underline{0.0336{\scriptsize$\pm$0.0062}} \\
    & & UNIStainNet & 101.7560 & 74.9900 & 0.5978{\scriptsize$\pm$0.0617} & 0.3037{\scriptsize$\pm$0.0796} & 10.7517{\scriptsize$\pm$2.2858} & 0.1845{\scriptsize$\pm$0.0781} & 0.1177{\scriptsize$\pm$0.0257} \\
    & & UNSB & 57.9643 & 20.5260 & 0.5692{\scriptsize$\pm$0.0640} & 0.2808{\scriptsize$\pm$0.0547} & 12.8627{\scriptsize$\pm$2.3306} & 0.2234{\scriptsize$\pm$0.0721} & 0.1002{\scriptsize$\pm$0.0307} \\
    & & \textbf{SheafStain (ours)} & \textbf{34.5080} & \textbf{2.2350} & \textbf{0.5196{\scriptsize$\pm$0.0646}} & \textbf{0.2056{\scriptsize$\pm$0.0382}} & \underline{14.1253{\scriptsize$\pm$2.7078}} & \underline{0.2336{\scriptsize$\pm$0.0861}} & \textbf{0.0292{\scriptsize$\pm$0.0080}} \\
    \cmidrule(lr){2-10}
    & \multirow{7}{*}{ER} & pix2pix & 208.5043 & 210.1640 & 0.5735{\scriptsize$\pm$0.0657} & 0.4237{\scriptsize$\pm$0.0815} & \textbf{15.3260{\scriptsize$\pm$3.3079}} & \textbf{0.2907{\scriptsize$\pm$0.0872}} & 0.0692{\scriptsize$\pm$0.0259} \\
    & & CycleGAN & 67.1467 & 28.0530 & 0.5492{\scriptsize$\pm$0.0549} & 0.2735{\scriptsize$\pm$0.0616} & 13.3093{\scriptsize$\pm$2.6881} & \underline{0.2633{\scriptsize$\pm$0.1170}} & 0.0693{\scriptsize$\pm$0.0228} \\
    & & PSPStain & \underline{45.9200} & 16.0910 & \underline{0.5235{\scriptsize$\pm$0.0496}} & 0.2534{\scriptsize$\pm$0.0493} & 13.9979{\scriptsize$\pm$2.7681} & 0.2535{\scriptsize$\pm$0.0868} & 0.0752{\scriptsize$\pm$0.0286} \\
    & & D-VST & 84.2247 & 65.3180 & 0.5518{\scriptsize$\pm$0.0384} & 0.2598{\scriptsize$\pm$0.0632} & 12.8127{\scriptsize$\pm$2.4215} & 0.2347{\scriptsize$\pm$0.0836} & \underline{0.0310{\scriptsize$\pm$0.0063}} \\
    & & UNIStainNet & 74.8386 & 37.2410 & 0.5259{\scriptsize$\pm$0.0550} & \underline{0.2461{\scriptsize$\pm$0.0541}} & 13.9829{\scriptsize$\pm$2.6328} & 0.2513{\scriptsize$\pm$0.0879} & 0.0571{\scriptsize$\pm$0.0112} \\
    & & UNSB & 49.2278 & \underline{16.0520} & 0.5659{\scriptsize$\pm$0.0545} & 0.2710{\scriptsize$\pm$0.0606} & 13.2345{\scriptsize$\pm$2.5429} & 0.2530{\scriptsize$\pm$0.0836} & 0.0924{\scriptsize$\pm$0.0279} \\
    & & \textbf{SheafStain (ours)} & \textbf{29.0824} & \textbf{2.5000} & \textbf{0.5042{\scriptsize$\pm$0.0619}} & \textbf{0.1934{\scriptsize$\pm$0.0358}} & \underline{14.1250{\scriptsize$\pm$2.9870}} & 0.2550{\scriptsize$\pm$0.0971} & \textbf{0.0292{\scriptsize$\pm$0.0075}} \\
    \cmidrule(lr){2-10}
    & \multirow{7}{*}{PR} & pix2pix & 210.9157 & 206.3250 & 0.5788{\scriptsize$\pm$0.0639} & 0.4148{\scriptsize$\pm$0.0754} & \textbf{15.6652{\scriptsize$\pm$3.6535}} & \textbf{0.3271{\scriptsize$\pm$0.1100}} & 0.0686{\scriptsize$\pm$0.0331} \\
    & & CycleGAN & 140.6118 & 99.8560 & 0.5783{\scriptsize$\pm$0.0474} & 0.3046{\scriptsize$\pm$0.0621} & 12.8598{\scriptsize$\pm$2.6051} & 0.2639{\scriptsize$\pm$0.1106} & 0.1022{\scriptsize$\pm$0.0238} \\
    & & PSPStain & \underline{48.6565} & \underline{19.2100} & 0.5447{\scriptsize$\pm$0.0463} & 0.2744{\scriptsize$\pm$0.0564} & 14.0482{\scriptsize$\pm$2.8649} & 0.2718{\scriptsize$\pm$0.0929} & 0.0797{\scriptsize$\pm$0.0287} \\
    & & D-VST & 90.1997 & 67.5810 & 0.5566{\scriptsize$\pm$0.0400} & \underline{0.2650{\scriptsize$\pm$0.0631}} & 12.9149{\scriptsize$\pm$2.6960} & 0.2437{\scriptsize$\pm$0.0869} & \underline{0.0308{\scriptsize$\pm$0.0064}} \\
    & & UNIStainNet & 91.2102 & 60.3800 & \underline{0.5232{\scriptsize$\pm$0.0729}} & 0.3060{\scriptsize$\pm$0.0916} & \underline{15.0155{\scriptsize$\pm$3.7727}} & \underline{0.3010{\scriptsize$\pm$0.0941}} & 0.0351{\scriptsize$\pm$0.0059} \\
    & & UNSB & 54.2625 & 21.3240 & 0.5741{\scriptsize$\pm$0.0594} & 0.2797{\scriptsize$\pm$0.0639} & 13.4669{\scriptsize$\pm$3.0705} & 0.2757{\scriptsize$\pm$0.0915} & 0.0904{\scriptsize$\pm$0.0368} \\
    & & \textbf{SheafStain (ours)} & \textbf{29.7240} & \textbf{2.1910} & \textbf{0.5017{\scriptsize$\pm$0.0588}} & \textbf{0.1962{\scriptsize$\pm$0.0361}} & 14.3530{\scriptsize$\pm$3.2782} & 0.2661{\scriptsize$\pm$0.1034} & \textbf{0.0272{\scriptsize$\pm$0.0075}} \\
    \cmidrule(lr){2-10}
    & \multirow{7}{*}{Ki-67} & pix2pix & 253.8902 & 306.2140 & 0.5707{\scriptsize$\pm$0.0371} & 0.3617{\scriptsize$\pm$0.0422} & \textbf{15.5375{\scriptsize$\pm$2.0329}} & \textbf{0.2949{\scriptsize$\pm$0.0824}} & 0.0839{\scriptsize$\pm$0.0180} \\
    & & CycleGAN & 74.4676 & 43.0280 & 0.5552{\scriptsize$\pm$0.0433} & 0.2575{\scriptsize$\pm$0.0408} & 13.8314{\scriptsize$\pm$1.9107} & 0.2714{\scriptsize$\pm$0.1062} & 0.0618{\scriptsize$\pm$0.0164} \\
    & & PSPStain & 44.0905 & 15.9350 & 0.5442{\scriptsize$\pm$0.0345} & 0.2618{\scriptsize$\pm$0.0383} & \underline{14.3419{\scriptsize$\pm$1.8375}} & 0.2778{\scriptsize$\pm$0.0824} & 0.0710{\scriptsize$\pm$0.0226} \\
    & & D-VST & 74.1956 & 54.2740 & 0.5587{\scriptsize$\pm$0.0324} & 0.2414{\scriptsize$\pm$0.0376} & 13.7817{\scriptsize$\pm$2.1425} & 0.2714{\scriptsize$\pm$0.0918} & \underline{0.0282{\scriptsize$\pm$0.0068}} \\
    & & UNIStainNet & 62.7007 & 37.8290 & \underline{0.5420{\scriptsize$\pm$0.0359}} & \underline{0.2254{\scriptsize$\pm$0.0361}} & 14.2793{\scriptsize$\pm$2.3256} & \underline{0.2879{\scriptsize$\pm$0.0970}} & 0.0620{\scriptsize$\pm$0.0209} \\
    & & UNSB & \underline{38.1956} & \underline{10.1780} & 0.5539{\scriptsize$\pm$0.0351} & 0.2514{\scriptsize$\pm$0.0397} & 13.7903{\scriptsize$\pm$1.9801} & 0.2740{\scriptsize$\pm$0.0872} & 0.0857{\scriptsize$\pm$0.0225} \\
    & & \textbf{SheafStain (ours)} & \textbf{25.3978} & \textbf{3.9220} & \textbf{0.5269{\scriptsize$\pm$0.0404}} & \textbf{0.2124{\scriptsize$\pm$0.0324}} & 14.2349{\scriptsize$\pm$2.0228} & 0.2771{\scriptsize$\pm$0.0987} & \textbf{0.0261{\scriptsize$\pm$0.0066}} \\
    \bottomrule
  \end{tabular}%
  }
\end{table}

\paragraph{Interpreting PSNR and SSIM under weak pairing.}
H\&E and IHC images are acquired from consecutive tissue sections rather than
the same physical slice (Section~\ref{sec:intro}), so pixel-exact correspondence
between input and target is fundamentally absent. Even same-section re-staining
protocols (e.g., HER2match~\citep{her2match}) incur tissue damage and registration
drift through the bleach-and-restain process, so pixel-aligned ground truth is
not recoverable in practice. PSNR and SSIM penalize
pixel-level intensity and local-structure deviation, favoring outputs regressed
toward a smoothed or averaged target. The pix2pix baseline illustrates this
trade-off. It attains the highest PSNR and SSIM in nearly every (dataset, stain)
block of Table~\ref{tab:full_metrics_1024_a} despite recording substantially
worse FID and KID, and ranking sixth of seven on DAB-$r$ for BCI HER2.
The metrics also differ in their sensitivity to the misalignment itself.
PSNR and SSIM are pixel-aligned and degrade under sub-cellular displacement,
the same mechanism that bounds DAB-$r$ above. DISTS is constructed to be
relatively insensitive to geometric transformations such as translation and
dilation~\citep{dists}, and LPIPS is likewise more tolerant, while FID and KID
use no pairing at all.
We therefore weight distribution-level
(FID, KID), perceptual (LPIPS, DISTS), stitching (TS), and biological
(DAB-$r$, DAB-KL, DAB-JSD, mIOD, FOD) metrics more heavily when interpreting
Tables~\ref{tab:full_metrics_1024_a} and~\ref{tab:full_metrics_1024_b}, since
these are less sensitive to the pixel-alignment artifact intrinsic to
weakly-paired data than PSNR and SSIM are. DAB-$r$ is the exception among them.
It measures spatial agreement directly, so the same misalignment bounds it from
above.

\begin{table}[h]
  \caption{Full biomarker quantification metrics at $1024 \times 1024$ (companion to Table~\ref{tab:full_metrics_1024_a}).
           DAB intensity correlation, KL/JSD divergences between DAB histograms,
           and absolute errors of mIOD and FOD. Same protocol convention as Table~\ref{tab:full_metrics_1024_a}.}
  \label{tab:full_metrics_1024_b}
  \centering
  \scriptsize
  \setlength{\tabcolsep}{3pt}
  \resizebox{0.90\textwidth}{!}{%
  \begin{tabular}{lll*{5}{>{\centering\arraybackslash}p{5.0em}}}
    \toprule
    Dataset & Stain & Method & DAB-$r$$\uparrow$ & DAB-KL$\downarrow$ & DAB-JSD$\downarrow$ & mIOD$\downarrow$ & FOD$\downarrow$ \\
    \midrule
    \multirow{7}{*}{BCI} & \multirow{7}{*}{HER2} & pix2pix & 0.0559 & \underline{0.8688} & \textbf{0.1085} & \underline{0.0838} & \textbf{0.0866} \\
    & & CycleGAN & 0.0668 & 2.7118 & 0.3006 & 0.1479 & 0.1954 \\
    & & PSPStain & 0.0317 & 1.9180 & 0.2276 & 0.1174 & 0.1679 \\
    & & D-VST & 0.0736 & \textbf{0.8404} & 0.1346 & 0.1172 & \underline{0.1003} \\
    & & UNIStainNet & \textbf{0.1267} & 1.1834 & 0.1591 & 0.0946 & 0.1280 \\
    & & UNSB & 0.0589 & 1.0434 & 0.2164 & 0.1674 & 0.2227 \\
    & & \textbf{SheafStain (ours)} & \underline{0.1209} & 1.2226 & \underline{0.1255} & \textbf{0.0615} & 0.1149 \\
    \midrule
    \multirow{28}{*}{MIST} & \multirow{7}{*}{HER2} & pix2pix & \underline{0.0866} & 1.2894 & 0.1286 & 0.2195 & 0.1877 \\
    & & CycleGAN & 0.0681 & 0.9302 & 0.1401 & 0.2525 & 0.2057 \\
    & & PSPStain & 0.0841 & 0.6306 & 0.1109 & \underline{0.1676} & 0.1964 \\
    & & D-VST & 0.0356 & \underline{0.4840} & \underline{0.1025} & 0.1964 & \underline{0.1837} \\
    & & UNIStainNet & 0.0606 & 1.2083 & 0.2222 & 0.4749 & 0.2613 \\
    & & UNSB & 0.0310 & 0.8462 & 0.1452 & 0.2648 & 0.2137 \\
    & & \textbf{SheafStain (ours)} & \textbf{0.1187} & \textbf{0.3447} & \textbf{0.0704} & \textbf{0.0714} & \textbf{0.1650} \\
    \cmidrule(lr){2-8}
    & \multirow{7}{*}{ER} & pix2pix & 0.0551 & 1.9522 & 0.1686 & 0.2991 & 0.2119 \\
    & & CycleGAN & 0.0884 & 0.8162 & 0.1386 & 0.2728 & 0.2355 \\
    & & PSPStain & \underline{0.0906} & 0.5176 & \underline{0.0999} & \underline{0.1963} & \underline{0.1852} \\
    & & D-VST & 0.0554 & \underline{0.4522} & 0.1072 & 0.2526 & 0.2226 \\
    & & UNIStainNet & 0.0497 & 1.2837 & 0.1288 & 0.2801 & 0.2170 \\
    & & UNSB & 0.0790 & 0.8176 & 0.1305 & 0.2698 & 0.2341 \\
    & & \textbf{SheafStain (ours)} & \textbf{0.1248} & \textbf{0.3890} & \textbf{0.0794} & \textbf{0.0794} & \textbf{0.1692} \\
    \cmidrule(lr){2-8}
    & \multirow{7}{*}{PR} & pix2pix & 0.0656 & 1.6081 & 0.1523 & 0.3132 & 0.1937 \\
    & & CycleGAN & 0.0823 & 0.6455 & 0.1347 & 0.2945 & 0.2219 \\
    & & PSPStain & \underline{0.0905} & \underline{0.4411} & \underline{0.0900} & \underline{0.2205} & \underline{0.1783} \\
    & & D-VST & 0.0502 & 0.4854 & 0.1170 & 0.2741 & 0.2229 \\
    & & UNIStainNet & 0.0429 & 3.0349 & 0.1885 & 0.3395 & 0.3417 \\
    & & UNSB & 0.0544 & 1.1273 & 0.1546 & 0.3181 & 0.2527 \\
    & & \textbf{SheafStain (ours)} & \textbf{0.1184} & \textbf{0.3572} & \textbf{0.0772} & \textbf{0.0954} & \textbf{0.1676} \\
    \cmidrule(lr){2-8}
    & \multirow{7}{*}{Ki-67} & pix2pix & 0.0257 & 0.7639 & 0.1202 & 0.1490 & \textbf{0.2067} \\
    & & CycleGAN & \underline{0.0423} & 0.7734 & 0.1323 & 0.1384 & 0.2412 \\
    & & PSPStain & 0.0340 & 0.5525 & 0.1052 & \underline{0.1135} & 0.2134 \\
    & & D-VST & 0.0235 & \textbf{0.4586} & \underline{0.1036} & 0.1267 & 0.2191 \\
    & & UNIStainNet & 0.0200 & 0.6261 & 0.1339 & 0.1163 & 0.2735 \\
    & & UNSB & 0.0326 & 0.6734 & 0.1299 & 0.1400 & 0.2585 \\
    & & \textbf{SheafStain (ours)} & \textbf{0.0493} & \underline{0.5217} & \textbf{0.1001} & \textbf{0.0750} & \underline{0.2072} \\
    \bottomrule
  \end{tabular}%
  }
\end{table}

\paragraph{Interpreting distribution-distance and absolute-error DAB metrics.}
DAB-KL, DAB-JSD, mIOD, and FOD reward predictions that statistically match
target chromogen distributions without placing the chromogen in the corresponding spatial regions, 
admitting the same regression-to-empty advantage as PSNR and SSIM on sparse-marker images. On BCI HER2, where many test images
are HER2 score 0/1+ with near-zero ground-truth DAB, pix2pix is first or
second on all four of these metrics while ranking sixth of seven on DAB-$r$
($0.0559$). Its histograms match the empty regions of the ground truth, but its
predictions do not place the chromogen where the reference has it. 
On MIST Ki-67, D-VST's two-stage tone-injection inference matches global DAB histograms (best DAB-KL) at $2.1\times$ lower DAB-$r$ than SheafStain. 
On MIST HER2, ER, and PR, SheafStain is best on all five of these metrics, so the caveat does not arise there. 
We therefore prioritize DAB-$r$, the spatial-correlation metric quantifying whether predicted DAB is placed in the same regions as the reference. 
SheafStain attains the highest DAB-$r$ on four of the five (dataset, stain) blocks of Table~\ref{tab:full_metrics_1024_b} and is second on BCI HER2.
HER2 binary classification (Table~\ref{tab:downstream_cls},
Appendix~\ref{app:downstream}) confirms the clinical relevance of DAB-$r$
alignment. SheafStain leads accuracy, F1, and AUROC on real BCI test data,
exceeding the strongest prior method (UNIStainNet) by $+4.2$\,pp accuracy
and $+0.057$ AUROC.

\paragraph{The attainable ceiling on DAB-$r$ under weak pairing.}
Because H\&E and IHC are cut from two different sections of the same tissue, the reference is
spatially and biologically offset from the input, which bounds DAB-$r$ for any
model. On the full evaluation splits we correlate each ground-truth DAB map
with a displaced copy of itself. At the BCI resolution of $0.46\,\mu$m/px the
attainable DAB-$r$ falls from $0.89$ at $0.9\,\mu$m to $0.57$ at $3.7\,\mu$m,
$0.40$ at $7.4\,\mu$m, $0.27$ at $14.7\,\mu$m, and $0.19$ at $29.4\,\mu$m.
A calibrated elastic warp of the same mean magnitude gives comparable values
($0.46$ at $7.4\,\mu$m and $0.33$ at $14.7\,\mu$m) and is insensitive to the
warp smoothness. Consecutive sections retain a median registration residual of
$6.5$ to $24.1\,\mu$m even after non-rigid registration~\citep{lotz2023}, so a
perfect but misaligned predictor is capped at roughly $0.21$ to $0.43$. The
metric is intrinsically compressed, since a displacement of one micron already
reduces it to about $0.89$. Observed values fall below even this geometric
bound because the two sections also differ biologically, which registration
cannot recover, and no method in Table~\ref{tab:full_metrics_1024_b} exceeds
about $0.13$. The bound applies to every method, so the comparison across
methods remains the quantity that carries information.

\FloatBarrier
\clearpage
\section{Qualitative Performance Comparison}
\label{app:qualitative}

\begin{figure}[h]
  \centering
  \includegraphics[width=0.9\linewidth]{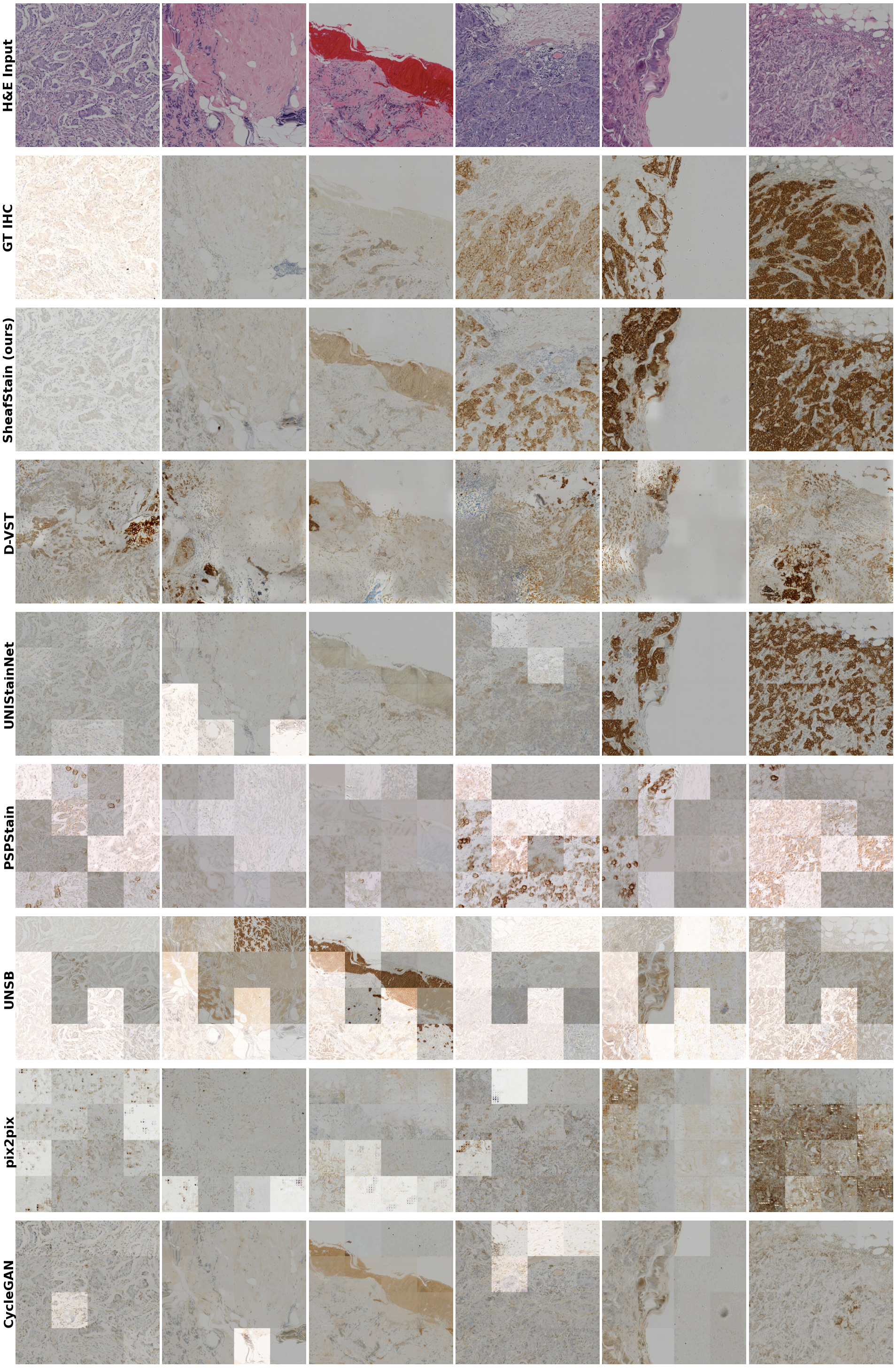}
  \caption{\textbf{Additional HER2 qualitative comparison on BCI at $1024 \times 1024$.}}
  \label{fig:app_qual_bci_her2}
\end{figure}

\begin{figure}[h]
  \centering
  \includegraphics[width=0.9\linewidth]{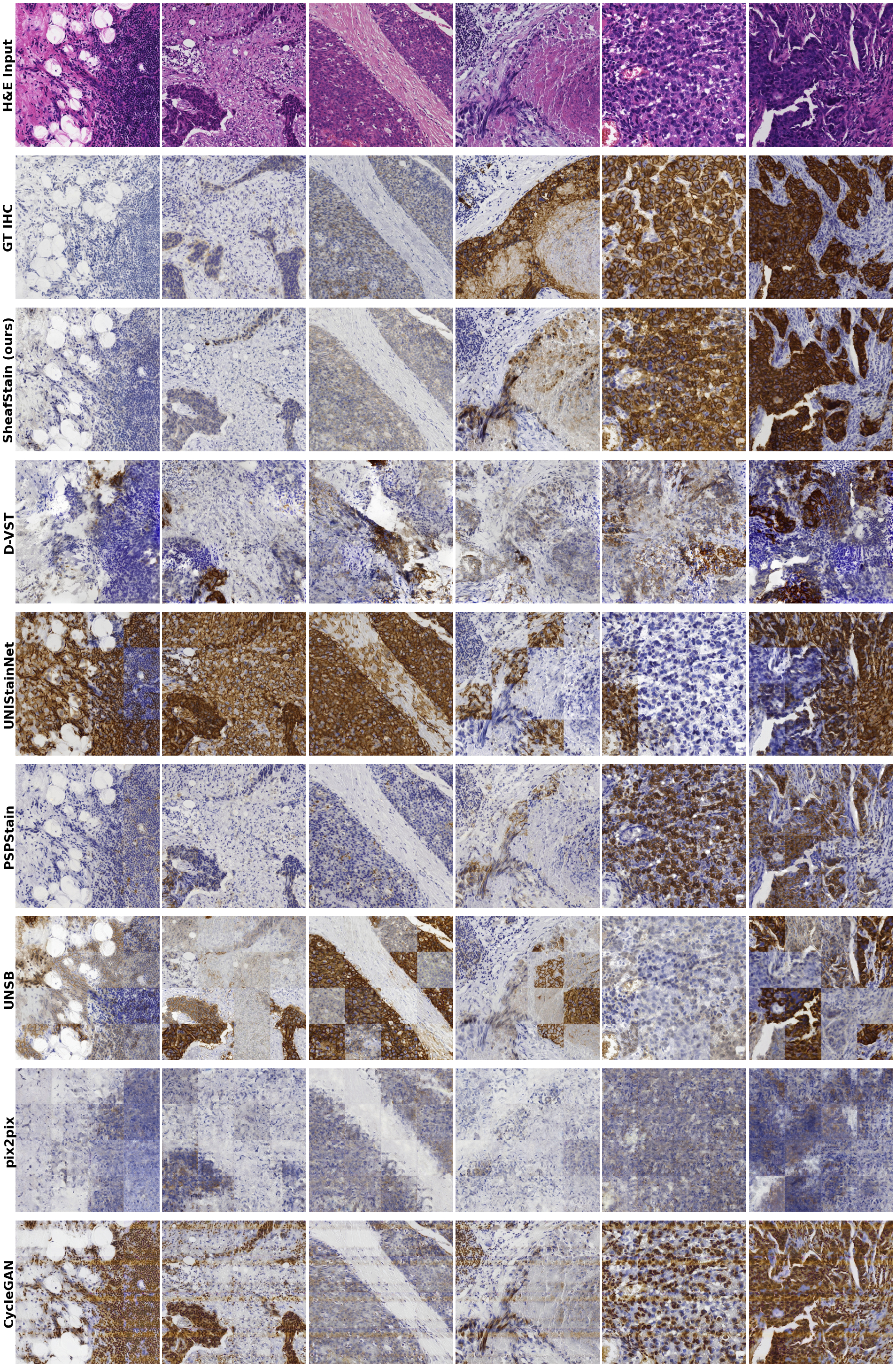}
  \caption{\textbf{Additional HER2 qualitative comparison on MIST at $1024 \times 1024$.}}
  \label{fig:app_qual_mist_her2}
\end{figure}

\begin{figure}[h]
  \centering
  \includegraphics[width=0.9\linewidth]{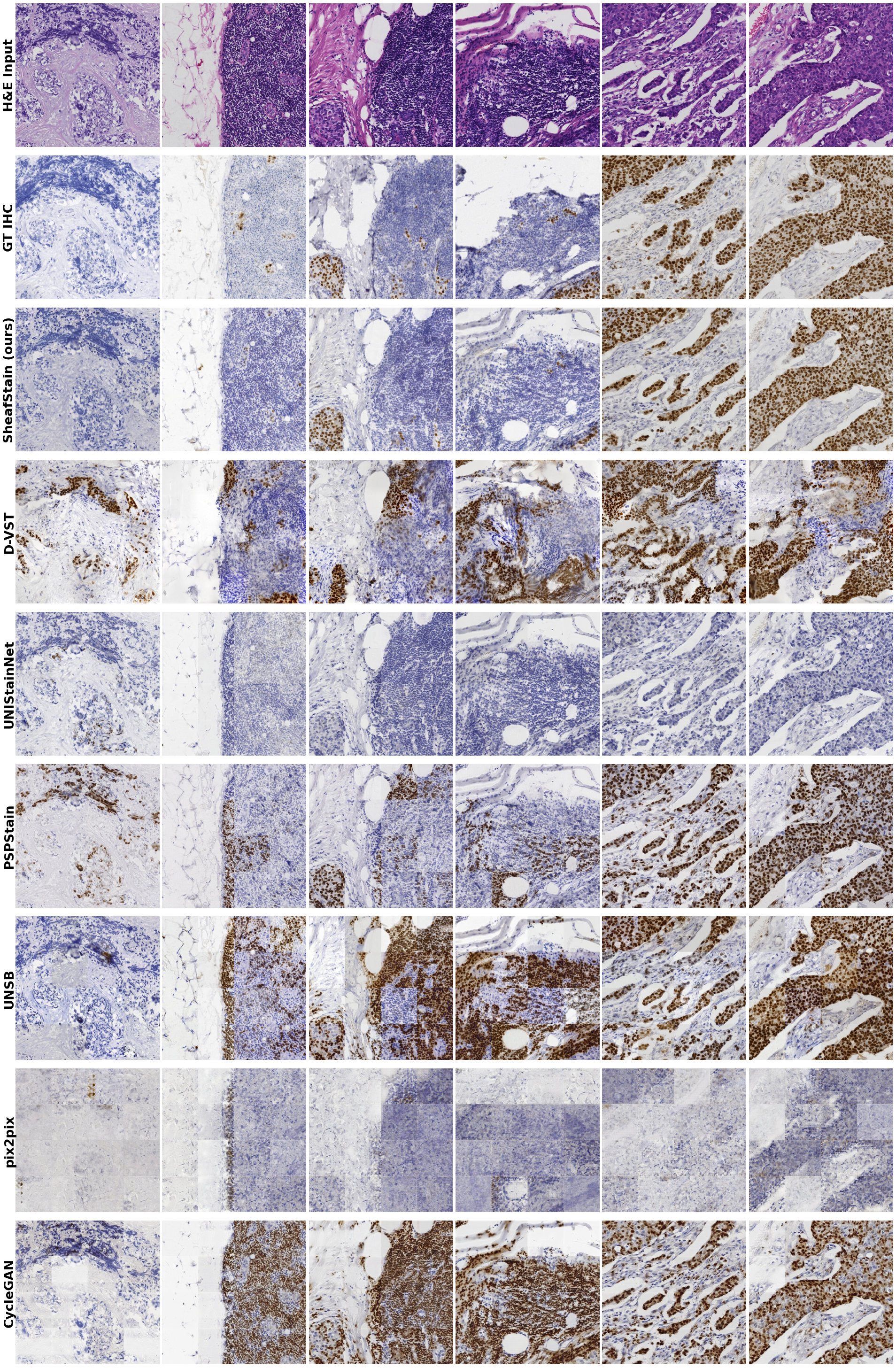}
  \caption{\textbf{ER qualitative comparison on MIST at $1024 \times 1024$.}}
  \label{fig:app_qual_mist_er}
\end{figure}

\begin{figure}[h]
  \centering
  \includegraphics[width=0.9\linewidth]{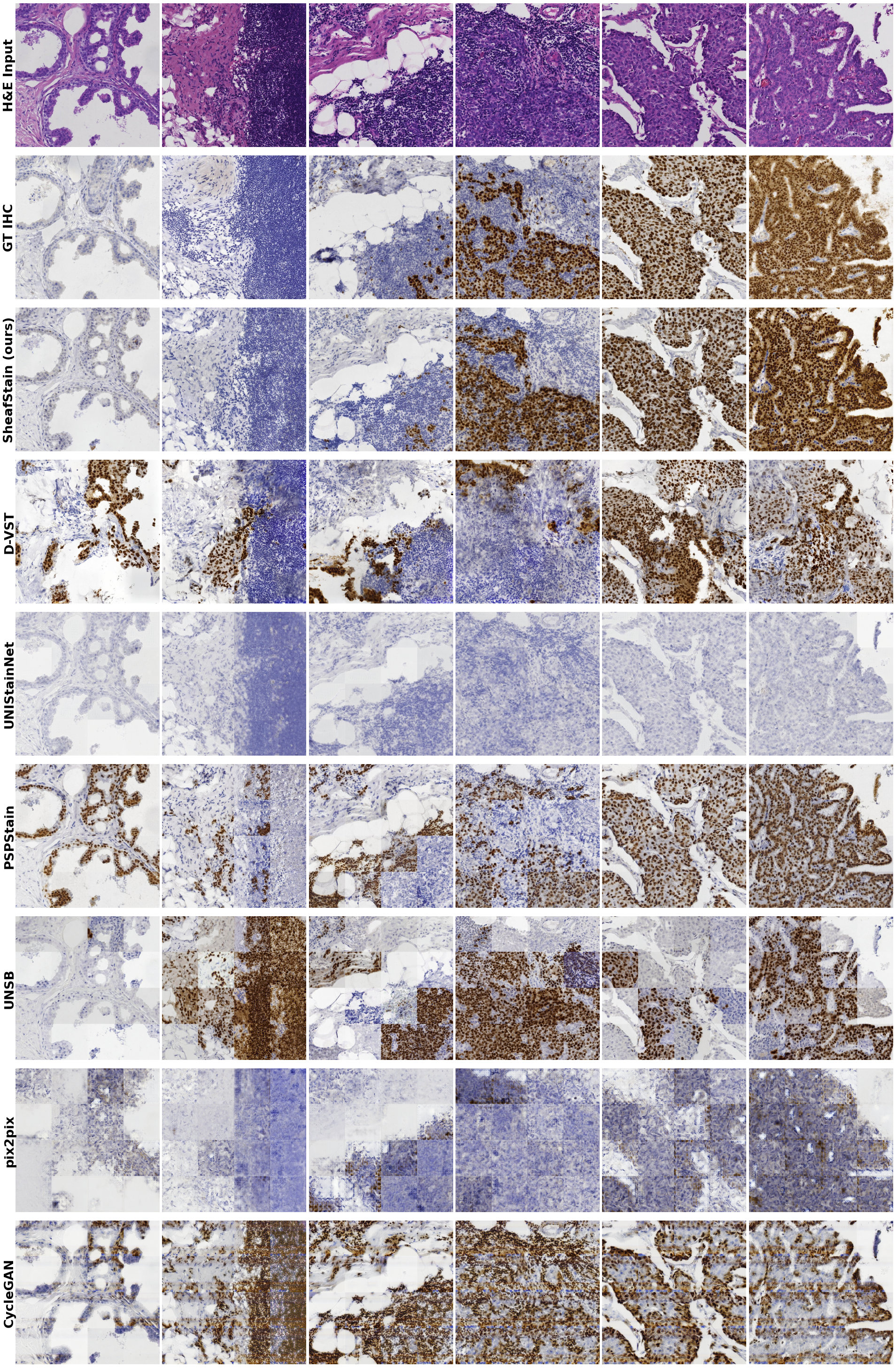}
  \caption{\textbf{PR qualitative comparison on MIST at $1024 \times 1024$.}}
  \label{fig:app_qual_mist_pr}
\end{figure}

\begin{figure}[h]
  \centering
  \includegraphics[width=0.9\linewidth]{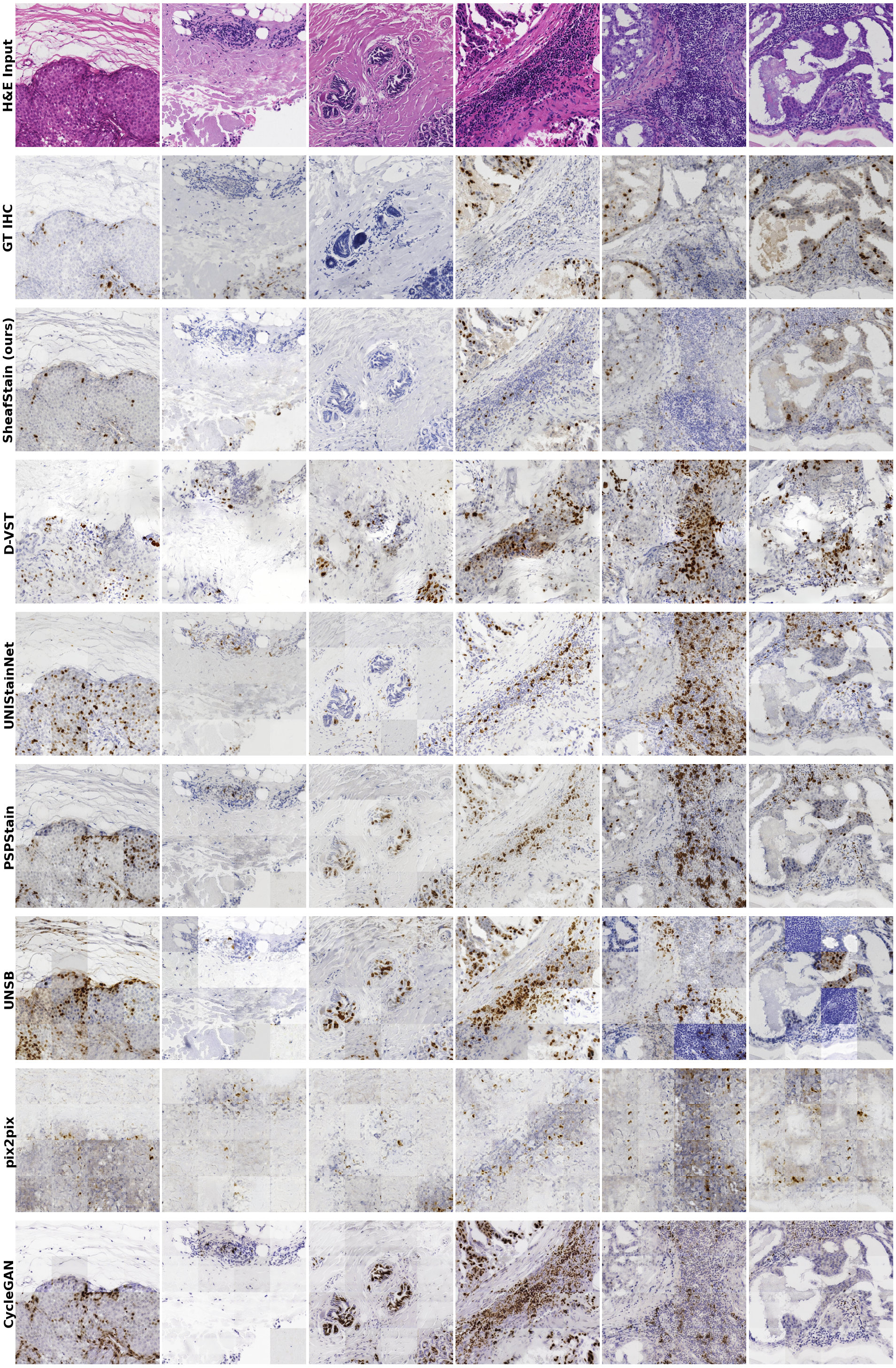}
  \caption{\textbf{Ki-67 qualitative comparison on MIST at $1024 \times 1024$.}}
  \label{fig:app_qual_mist_ki67}
\end{figure}

This section complements Figure~\ref{fig:qualitative} along two axes.
Figures~\ref{fig:app_qual_bci_her2} and~\ref{fig:app_qual_mist_her2} broaden the HER2 comparison on the BCI and MIST datasets
beyond the two cases per dataset shown in Figure~\ref{fig:qualitative}.
Figures~\ref{fig:app_qual_mist_er},~\ref{fig:app_qual_mist_pr}, and~\ref{fig:app_qual_mist_ki67} extend the comparison from HER2 to the
remaining MIST stains (ER, PR, and Ki-67).

\FloatBarrier
\section{HER2 Binary Classification Details}
\label{app:downstream}

\subsection{Protocol and Rationale}
\label{appx:downstream:protocol}

BCI labels each image with one of the four standard HER2 IHC grades ($0$, $1+$, $2+$, $3+$).
Under ASCO/CAP reporting~\citep{wolff2018asco,wolff2023asco}, $0$ and $1+$ are negative, $2+$ is
equivocal and requires reflex in situ hybridization (ISH), and $3+$ is positive. We collapse the four
grades into Absent-to-Low (HER2-AL; $0, 1+$) and Equivocal-to-Positive (HER2-EP; $2+, 3+$). The split
thus follows the guideline's division between negative results that close HER2 testing when no
histopathologic discordance is apparent, and equivocal or positive results that proceed to reflex ISH
or a positive report. This is the decision a screening tool would support. The split also keeps the
$0$ vs.\ $1+$ and $2+$ vs.\ $3+$ distinctions, which show limited agreement among
pathologists~\citep{fernandez2022her2low}, inside a class rather than at the class boundary. In the
BCI test partition it yields $273$ HER2-AL and $704$ HER2-EP images, for which F1 and AUROC capture
both balanced discrimination and threshold-independent ranking quality. When discordance is suspected
(\eg, a negative core biopsy from a grade~3 tumor), the guideline allows a new HER2
test~\citep{wolff2018asco}.

Finer clinical categories are out of reach on BCI. Eligibility for trastuzumab deruxtecan now
depends on distinctions at the low end of the scale, between IHC $0$ and $1+$~\citep{modi2022her2low}
and, in hormone receptor-positive disease, within IHC $0$, where HER2-ultralow denotes faint
membrane staining in at most $10\%$ of tumor cells~\citep{bardia2024destinybreast06}. BCI provides
neither sub-categories of IHC $0$ nor ISH results for $2+$ cases. The $0$ vs.\ $1+$ distinction is also
the least reproducible. Among $18$ pathologists, at least $17$ agreed for only $24$ of the $92$ cases that
any reader scored $0$ ($26\%$), against $26$ of the $45$ cases that any reader scored $3+$
($58\%$)~\citep{fernandez2022her2low}.

For each generative method, we train a ResNet-50 classifier from ImageNet
initialization on the method's translated outputs of the official BCI train
split ($3{,}896$ images) at $1024 \times 1024$, batch size $16$,
learning rate $10^{-4}$ for $100$ epochs, and evaluate on the real BCI test set
($977$ images). A real-IHC ceiling classifier trained on the same protocol but on
real BCI IHC achieves accuracy $0.974$, F1 $0.982$, and AUROC $0.997$, indicating
that the classifier architecture itself is not the bottleneck.

We adopt a transfer protocol (training on translated outputs and evaluating on
real IHC) rather than the more common train-on-real / test-on-translated
direction. Training on synthetic and testing on real is a stricter probe of
whether translated outputs preserve clinically discriminative features. Any
features that are artifacts of the generator (e.g., aliasing, fixed color casts,
or scanner-specific texture) fail to transfer to real IHC, while features that
encode HER2 grade do transfer regardless of how closely the synthetic
outputs match real-IHC texture statistics. The opposite direction is confounded
by domain shift, so a method that produces texturally faithful but clinically
uninformative outputs can still score well.

\subsection{ROC Analysis}
\label{appx:downstream:roc}

\begin{figure}[h]
  \centering
  \includegraphics[width=0.85\linewidth]{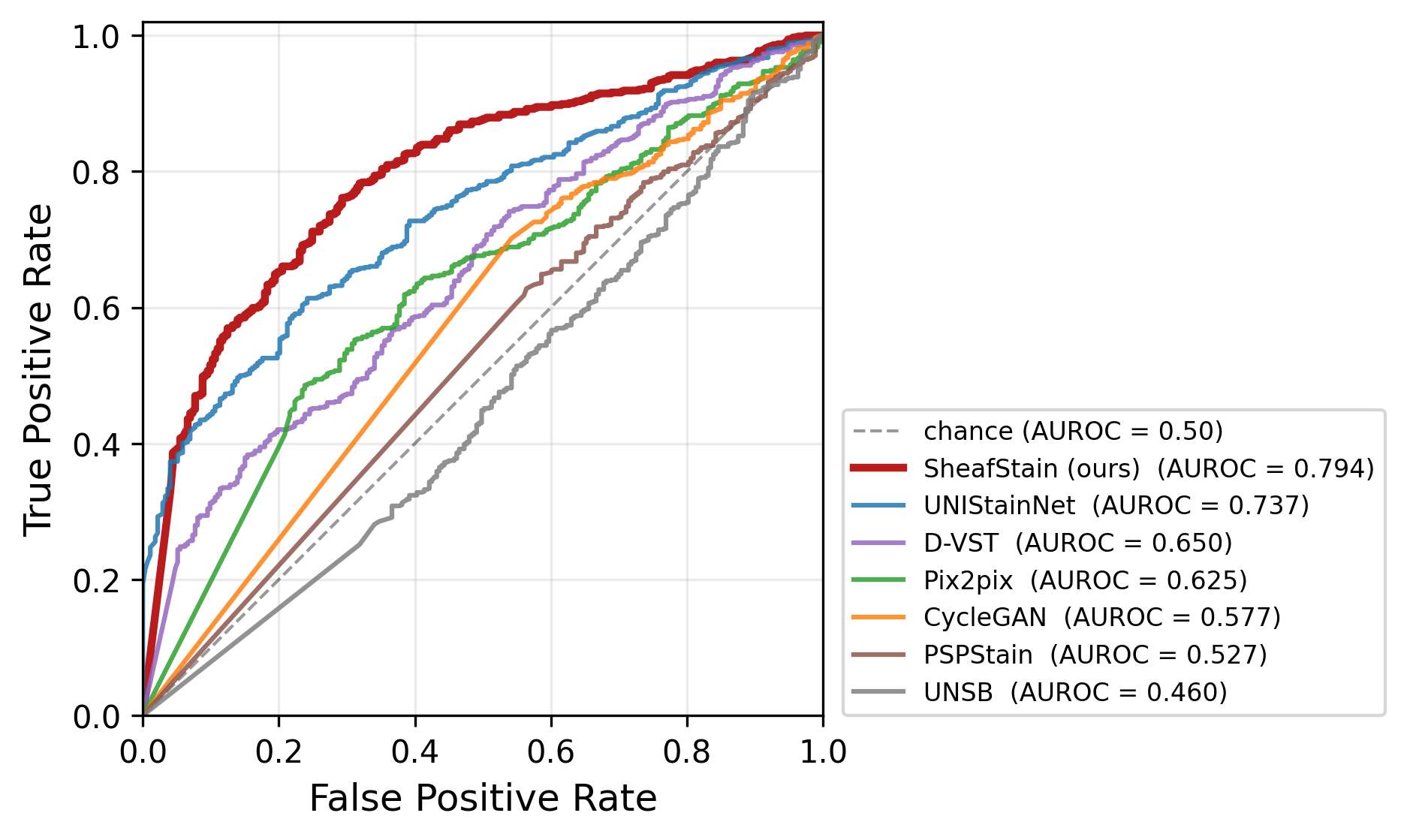}
  \caption{ROC curves on real BCI test for HER2-AL vs.\ HER2-EP classifiers, each
  trained on a different generative method's translated training and validation
  outputs. SheafStain attains the highest AUROC at $0.794$.}
  \label{fig:app_tstr_roc}
\end{figure}

Figure~\ref{fig:app_tstr_roc} overlays the ROC curves of all seven methods.
SheafStain reaches AUROC $0.794$, $0.057$ above the strongest
prior model (UNIStainNet, $0.737$). Three prior methods perform poorly on AUROC, most with raw accuracies in the
$0.65$--$0.71$ range: CycleGAN ($0.577$), PSPStain ($0.527$), and UNSB ($0.460$,
below chance), with UNSB the poorest of all seven methods.
This last comparison is informative because UNSB is the unpaired
Schr\"odinger bridge that SheafStain takes as its starting point. The $+0.334$ AUROC gap
from UNSB to SheafStain on the same underlying generator architecture isolates
the contribution of the sheaf-theoretic regularizers, VFM-derived spatial
conditioning, and the corresponding training signal introduced in this work
to downstream HER2-grade signal preservation.

\FloatBarrier

\section{Computational Cost Analysis}
\label{app:compute_cost}

We characterize where \method{} sits on the latency--quality trade-off curve
relative to six prior methods on the MIST HER2 validation set ($1024 \times 1024$, $N{=}1{,}000$).
\method{} occupies the high-quality endpoint of the non-dominated Pareto frontier
on both FID and DAB-$r$ (Section~\ref{appx:compcost:pareto}).

\subsection{Measurement Protocol}
\label{appx:compcost:protocol}

All methods are timed on a single GPU under identical measurement conditions
(hardware specifics: Section~\ref{appx:compcost:caveats}). Each method's production
checkpoint and inference protocol is wrapped as a callable that
maps a synthetic $1 \times 3 \times 1024 \times 1024$ tensor in
$[-1, 1]$ to its output, isolating GPU-side compute from disk
I/O, data pre-processing, and CLI overhead. We run $5$ warm-up
iterations followed by $N{=}100$ timed iterations, synchronizing
CUDA streams between iterations and resetting peak-memory
statistics after warm-up. Each method runs in its production
precision and uses PyTorch's default attention path. The
resulting standard deviation is below $1\%$ of the mean for
every method. Quality metrics (FID, KID, DISTS, and DAB-$r$) are computed separately on the full evaluation split.

\subsection{Latency--Quality Pareto Position}
\label{appx:compcost:pareto}

\begin{figure}[t]
\centering
\includegraphics[trim={0 0 0 43.6bp},clip,width=\linewidth]{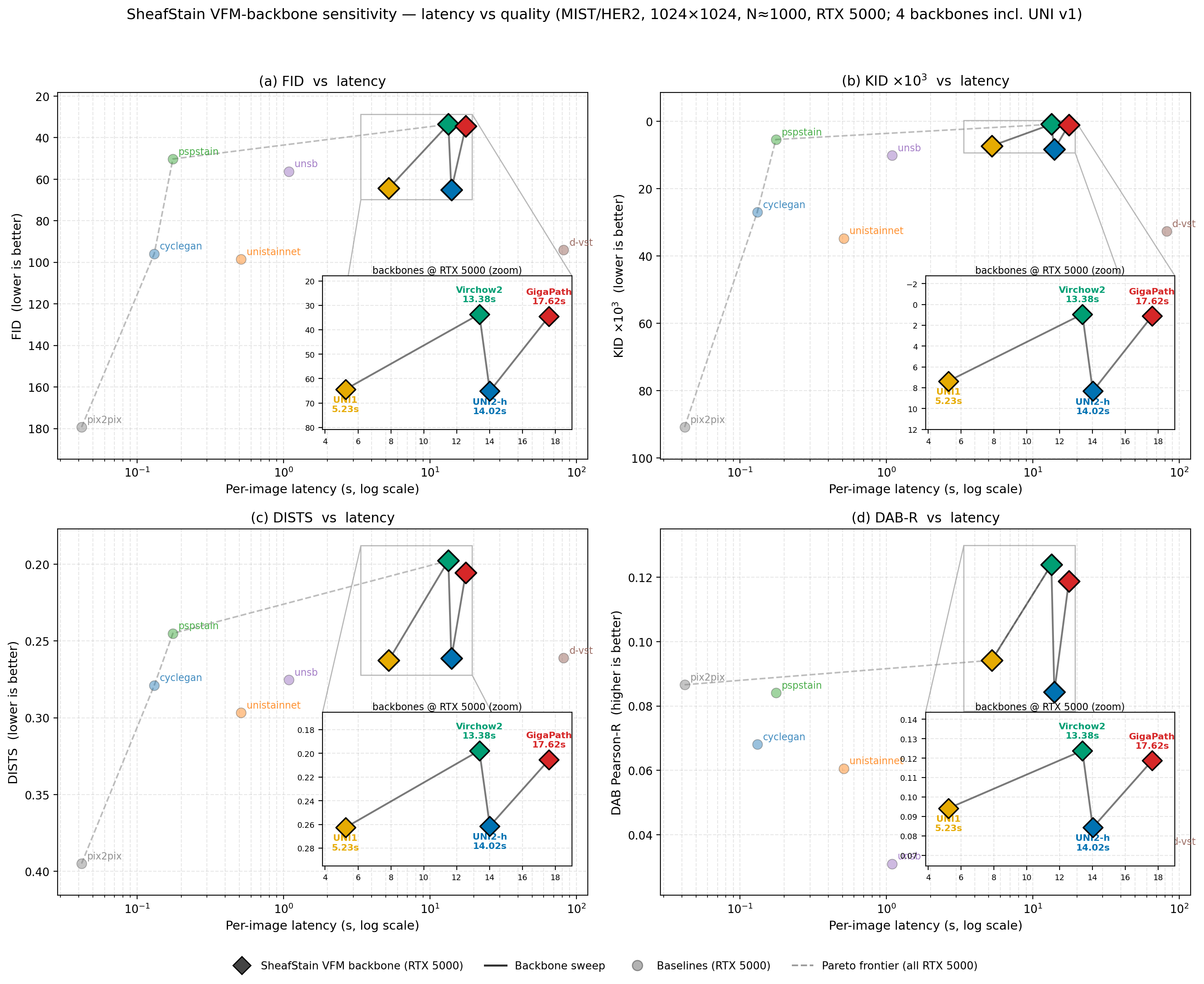}
\caption{\textbf{Latency--quality trade-off on MIST/HER2 across four VFM encoders.}
Measured on the MIST HER2 validation set ($1{,}000$ images) at $1024 \times 1024$,
on a single Quadro RTX~5000.
Four quality metrics vs.\ per-image latency: (a) FID, (b) KID$\times 10^{3}$,
(c) DISTS, and (d) DAB-$r$. All panels are oriented so that ``up is better''.
The y-axis is inverted for the lower-is-better metrics (a--c). Panel (d) uses a
normal y-axis for higher-is-better DAB-$r$. Methods on the dashed curve form
the non-dominated Pareto frontier (no other method is simultaneously faster
and higher-quality). Points off the curve are strictly worse than at least
one frontier method on both axes. Diamonds are \method{} with four
interchangeable VFM encoders (Prov-GigaPath, UNI~v1, UNI2-h, Virchow2).
Circles are the six prior methods. Insets zoom on the encoder sweep.
\method{} occupies the high-quality endpoint
on all four metrics.}
\label{fig:appx_pareto}
\end{figure}

The figure compares six prior methods against \method{} run with four
interchangeable VFM encoders. These ten configurations span four orders of magnitude in per-image
latency, from $\sim\!42$\,ms (Pix2pix) to $\sim\!82$\,s (D-VST),
so a logarithmic latency axis is essential for visual
comparison. Pix2pix holds the low-cost endpoint of the non-dominated Pareto
frontier (dashed curve in Figure~\ref{fig:appx_pareto}) and \method{} the
high-quality endpoint on all four metrics. PSPStain lies on the frontier for
FID, KID, and DISTS and falls off it for DAB-$r$, where Pix2pix is both faster
and higher ($0.087$ versus $0.084$). \method{} attains the
best FID ($34.5$, a $31\%$ relative reduction versus PSPStain
at $50.1$) and the best DAB-$r$ ($0.119$, $41\%$ higher
than PSPStain) at $\sim\!18$\,s/image of GPU compute, and exchanging its
VFM encoder for Virchow2~\citep{virchow} reaches comparable quality at about $13$\,s/image. CycleGAN also lies on the frontier for FID, KID, and DISTS and falls off it for
DAB-$r$. The remaining three prior methods (UNIStainNet, UNSB, D-VST)
are dominated on both axes simultaneously. In particular,
D-VST's $25$-step DPM-Solver++ diffusion pipeline is
$\sim\!4.7\times$ slower than \method{} and attains worse FID
($94.1$) and DAB-$r$ ($0.036$), placing it strictly below the
frontier.

\subsection{Per-Patch Vision Foundation Model Forward-Pass Accounting}
\label{appx:compcost:vfm}

The dominant component of \method{}'s per-image cost is the
VFM forward, about $90\%$ of the wall-clock cost,
used to extract the per-patch
\emph{neighborhood} feature map and CLS token that condition our
sheaf-aware generator. We use Prov-GigaPath ViT-G (1.13~B parameters) 
at $224 \times 224$ input resolution. 
A single VFM forward yields $223.45$~GFLOPs ($111.72$~GMACs), 
measured with \texttt{fvcore}'s analytical flop counter directly 
on the \texttt{forward\_features} graph. Per $1024 \times 1024$ image,
our inference protocol (stride $192$, $5{\times}5{=}25$ reference
patches, $8$ cardinal/diagonal overlapping patches per reference) issues
$200$ VFM forwards organized into batches of $32$, totaling
$\mathbf{44.7~TFLOPs}$ of VFM compute per image. These figures exclude the $9$ additional
reference patches of Appendix~\ref{app:inference}. The conditional
generator (a 9-block ResNet with time-conditioned modulation)
contributes the remaining $\sim\!10\%$ of the wall-clock cost
($25$ patch forwards at $256 \times 256$).

For comparison, UNIStainNet's VFM (UNI~\citep{uni} ViT-L,
$304$~M parameters) operates at the global image level. A single
batched forward over $16$ sub-crops yields a $16 \times 16$
patch-token grid that is broadcast as conditioning to every
$256 \times 256$ patch. This design is $46\times$ cheaper in VFM
FLOPs than ours by virtue of two simultaneous savings:
$3.7\times$ from the smaller encoder (ViT-L vs.\ ViT-G) and
$12.5\times$ from a global rather than per-patch neighborhood
conditioning. The per-patch design is, however, a structural
prerequisite for the sheaf-coboundary objective in our generator,
which assigns each patch its own local context and enforces
consistency with overlapping neighbors. Replacing per-patch
neighborhoods with a global feature would forfeit the inductive
bias that drives our quality gains.

\subsection{Hardware Caveats}
\label{appx:compcost:caveats}

We measure on a single workstation-grade Quadro RTX~5000
(Turing, $11.2$~TFLOPs fp32, $89$~TFLOPs fp16). The absolute
latencies reported here are therefore an upper bound on
those attainable with data-center hardware. On A100/H100-class accelerators with
native fp16/bf16 tensor cores, ViT-G inference (the dominant
component of our cost) runs roughly $5{-}10\times$ faster, so
\method{} would project to $\sim\!2{-}3$\,s per image under
the same protocol. Because \emph{all} methods are measured on
the same hardware in this work, the relative ordering and the
trade-off structure of Figure~\ref{fig:appx_pareto} are
unaffected by this choice.

\subsection{Discussion}
\label{appx:compcost:discussion}

\method{} explicitly trades inference compute for quality. The
$\approx\!100\times$ latency overhead versus the strongest quality-comparable prior method (PSPStain) and the $\approx\!16\times$
overhead versus UNSB are concentrated almost entirely in the VFM forward, which is the same architectural choice
that yields our quality gains. That choice is a pathology-specialized ViT-G encoder evaluated per local neighborhood rather than at the global image level. 
We view this trade-off as a deliberate Pareto-frontier move toward higher-quality virtual staining rather than as evidence of a tunable knob. 
Globally pooled conditioning forfeits the per-patch neighborhoods the sheaf objective is defined on, so it is ruled out by the design. 
Shrinking the encoder is a softer constraint. Of the three smaller encoders in Figure~\ref{fig:appx_pareto}, UNI ViT-L
and UNI2-h~\citep{uni} roughly double the FID and give up $21$ to $29\%$ of DAB-$r$, while Virchow2 matches Prov-GigaPath on both at about three quarters of the latency.
Encoder quality is therefore the operative variable rather than parameter count, and whether a smaller encoder is a viable alternative depends on the encoder itself rather than its size.

\FloatBarrier

\end{document}